\documentclass[a4paper]{article} 
\usepackage{times}

\usepackage{hyperref}
\usepackage{url}

\usepackage[utf8]{inputenc} 
\usepackage[T1]{fontenc}    
\usepackage{hyperref}       
\usepackage{url}            
\usepackage{booktabs}       
\usepackage{amsfonts}       
\usepackage{nicefrac}       
\usepackage{microtype}      
\usepackage{natbib}
\usepackage{hyphenat}       
\usepackage[normalem]{ulem}

\usepackage{graphicx}
\usepackage{subcaption} 
\usepackage{color}
\usepackage{xspace}

\definecolor{bblue}{rgb}{0.125,0.121,0.874}
\newcommand{\blue}{\textcolor{bblue}{\textbf{blue}}\xspace}

\definecolor{ggreen}{rgb}{0.172,0.47,0.188}
\newcommand{\green}{\textcolor{ggreen}{\textbf{green}}\xspace}

\definecolor{rred}{rgb}{0.658,0.227,0.231}
\newcommand{\red}{\textcolor{rred}{\textbf{red}}\xspace}

\definecolor{ggrey}{rgb}{0.43,0.43,0.43}
\newcommand{\grey}{\textcolor{ggrey}{\textbf{grey}}\xspace}

\definecolor{graycolor}{rgb}{0.4,0.4,0.4}

\newcommand{\layer}[1]{#1}
\newcommand{\unit}[1]{#1}

\newcommand{\precision}{$precision$}

\begin{document}

\title{Are there any `object detectors' in the hidden layers of CNNs trained to identify objects or scenes?}

\author{Ella M. Gale\footnote{ella.gale@bristol.ac.uk} \and Nicholas Martin\footnote{nm13850@bristol.ac.uk} \and Ryan Blything\footnote{ryan.blything@bristol.ac.uk} \and Anh Nguyen\footnote{anhnguyen@auburn.edu} \and Jeffrey S. Bowers\footnote{j.bowers@bristol.ac.uk} \footnote{E.M.G., N.M., R.B. and J.B.: School of Psychological Science, University of Bristol, 12a Priory Road, Bristol BS8 1TU, UK. A.N.: Department of Computer Science and Software Engineering, Auburn University, AL, USA}}


\maketitle

\begin{abstract}
    Various methods of measuring unit selectivity have been developed with the aim of better understanding how neural networks work.  But the different measures provide divergent estimates of selectivity, and this has led to different conclusions regarding the conditions in which selective object representations are learned and the functional relevance of these representations. In an attempt to better characterize object selectivity, we undertake a comparison of various selectivity measures on a large set of units in AlexNet, including localist selectivity, precision, class-conditional mean activity selectivity (CCMAS), 
    the human interpretation of activation maximization (AM) images, and standard signal-detection measures.  We find that the different measures provide different estimates of object selectivity, with \precision and CCMAS measures providing misleadingly high estimates. Indeed, the most selective units had a poor hit-rate or a high false-alarm rate (or both) in object classification, making them poor object detectors.  We fail to find any units that are even remotely as selective as the `grandmother cell' units reported in recurrent neural networks. In order to generalize these results, we compared selectivity measures on units in VGG-16 and GoogLeNet trained on the ImageNet or Places-365 datasets that have been described as `object detectors'. Again, we find poor hit-rates and high false-alarm rates for object classification. We conclude that signal-detection measures provide a better assessment of single-unit selectivity compared to common alternative approaches, and that deep convolutional networks of image classification do not learn object detectors in their hidden layers.
    
\end{abstract}

\section{Introduction}

There is a long history of single-cell neurophysiological studies designed to characterize the response of single neurons to visual stimuli \citep[for reviews see ][]{J17, Bowers2019, quiroga2016neuronal}. A key finding is that neurons often respond to visual information in a highly selective manner, with cells in V1 responding selectively to simple visual stimuli, and cells in IT and perirhinal cortex responding selectively to high level visual information.  This has led to the so-called “standard model” that includes a hierarchy of visual neurons with neurons in the higher layers encoding more and more complex visual features \citep{riesenhuber2002visual}. Whether individual neurons selectively encode whole objects (localist representations or so-called “grandmother cells” is contentious \citep[see debate between][]{J18, Bowers2010, Plaut2010, QuianQuiroga2010}, but it is clear that single neurons can encode high level visual features in a highly selective manner.

Deep convolutional neural networks (DCNNs) trained to perform image classification \cite{106} are roughly designed around the architecture of the human visual system responsible for object recognition, and these models have been described as good theories of object recognition.  For example, \citet{kubilius2018cornet} wrote: \emph{``Deep artificial neural networks with spatially repeated processing (a.k.a., deep convolutional [Artificial Neural Networks]) have been established as the best class of candidate models of visual processing in primate ventral visual processing stream''} (p.1).  
Apart from the impressive success in identifying photographs of objects, researchers have claimed that the patterns of activation of units in these networks match the patterns of activations of neurons in various brain areas involved in object identification, as measured through Representational Similarity Analyses \citep{yamins2014performance}. These analyses do not consider the activations of single units, but rather, the similarities amongst patterns of activations in brains in DCNNs.

Recently there has been growing interest in analysing the activations of single units in DCNNs. A key advantage of working with artificial networks is that it is possible to systematically analyse all the units, and it is possible to present networks with a much larger set of images. Indeed, it is possible to assess the response of all units to all training-set images and characterize unit selectivity under these ideal conditions \cite{yosinski2015understanding,zeiler2014visualizing}. Nevertheless, just as in the case with neurons in visual cortex, there are disagreements about the degree of selectivity of units in DCNNs, with some researchers reporting that some networks learn “grandmother cells” \citep[e.g., ][]{1, lakretz2019emergence}, others claiming that the learned selective representations constitute ``object detectors" but not grandmother cells \citep[e.g., ][]{102}, and still others emphasizing the distributed nature of learned representations \citep{leavitt2020selectivity}. The different conclusions may be the byproduct of researchers studying different network architectures, or studying networks trained on different tasks, or using different selectivity measures that are not comparable.



\section{Background Research} 
In one line of research, \citet{1,6} assessed the selectivity of single hidden units in recurrent neural networks (RNNs) designed to model human short-term memory.  They reported many localist or `grandmother cell' units that were 100\% selective for specific letters or words, where all members of the selective category were more active than and disjoint from all non-members, as can be shown in jitterplots \citep{10} (see Fig.~\ref{fig:BowersEg}). A jitterplot depicts that activation of a single unit in response to multiple different inputs, with each point or label corresponding to a given input. For example, in Figure 1, the location of each labeled word along the x-axis indicates the unit's level of response to this word, with words assigned an arbitrary value along the y-axis to minimize overlap. The jitterplot on the left depicts a selective unit (for the letter `j'), and the jitterplot on the right is for a non-selective unit.

The authors argued that the recurrent network learned localist representations in order to co-activate multiple letters or words at the same time in short-term memory without producing ambiguous blends of overlapping distributed patterns (the so-called `superposition catastrophe'). Consistent with this hypothesis, localist units only emerged when the recurrent model was trained to recall a series of words (a condition in which the model needed to solve the superposition catastrophe), but did not emerge when the model was trained on letters or words one at a time \citep{1}. 

In parallel, researchers have reported selective units in the hidden layers of various CNNs trained to classify images into one of multiple categories  \citep{105,101,73,74}, for a review see \citet{J17}. For example, \citet{105} assessed the selectivity of units in hidden layers of two CNNs trained to classify over one million images into 1000 objects and 205 scene categories, respectively. They reported many highly selective units that they characterized as `object detectors' (as defined below) in both networks. Similarly, \citet{101} reported that CNNs trained on two different image datasets learned many highly selective hidden units based on a Class-Conditional Mean Activation Selectivity (CCMAS) measure (defined below).  
Instead of harnessing training-set images, \citet{28} generated preferred images that maximally activated hidden units in CNNs using Activation Maximization (we describe one version of Activation Maximization below) and observed that some of the images were interpretable. For example, as illustrated in the top right of Figure 1, a generated image that maximally activated one hidden unit looks like a lighthouse, consistent with the hypothesis that the unit selectively codes for this category. These later findings appear to be inconsistent with \citet{6} who failed to observe selective representations in fully connected NNs trained on stimuli one at a time, but again different networks were used, the models were trained on different tasks, and most importantly for present purposes, different measures of selectivity were used, and accordingly, it is difficult to directly compare results.

A better understanding of the relation between selectivity measures is vital given that different measures are frequently used to address similar issues.  For example, both the human interpretability of generated images \citep{le2013building} and localist selectivity \citep{1} have been used to make claims about ‘grandmother cells’, but it is not clear whether these two measures provide similar insights into unit selectivity.  Similarly, based on their \precision metric, \citet{105} claim that the object detectors learned in CNNs play an important role in identifying specific objects, whereas \citet{101} challenge this conclusion based on their finding that units with high CCMAS measures were not especially important in the performance of their CNNs and concluded: \emph{``...it implies that methods for understanding neural networks based on analyzing highly selective single units, or finding optimal inputs for single units, such as activation maximization \citep{74} may be misleading''}. This makes a direct comparison between selectivity measures all the more important.

\begin{figure}[t]
	\includegraphics[width=340px]{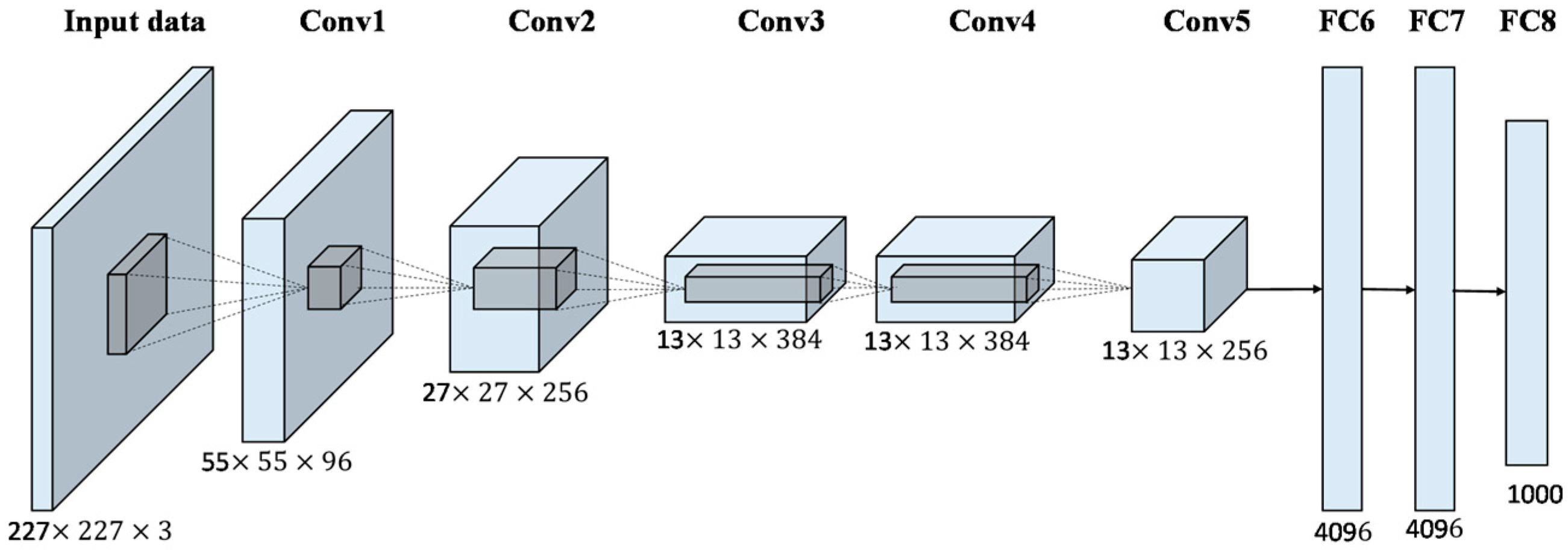}\\
	\caption{
    An illustration of the architecture of AlexNet, reprinted from \citep{han2017_alexNet_figure} with permission. The input layer is composed of 224$\times$224 units or the `retina' (far left) that encodes images and feeds the activated visual pixels into the first convolutional layer (\layer{conv1}).  \layer{conv1} learns 11$\times$11 filters (a.k.a. features) that are repeated across the input every four pixels (a stride of 4).  In \layer{conv1}, there are 96 different filter banks that each code for a different feature in the input across multiple retinal locations, much like there are different simple cells that encode different line orientations across multiple retinal locations. Different convolutional layers have different size filters and different number of filter banks, with `max pooling' layers (not depicted) occurring after the \layer{conv1}, \layer{conv2}, and \layer{conv5} layers.  The output of \layer{conv5} is then fed into a series of three fully connected (\layer{fc}) layers, with layers \layer{fc6} and \layer{fc7} each including 4096 units, and \layer{fc8} including 1000 units.  Each unit in \layer{fc8} codes for a single category. A softmax function is applied at \layer{fc8} to give the output probabilities for each learned category, in a localist or `one hot' encoding scheme.  We recorded from units in \layer{conv5}, \layer{fc6} and \layer{fc8}. 
	}
	\label{fig:AlexNetArch}
\end{figure}

\begin{figure}[t]
    
	\includegraphics[width=131px]{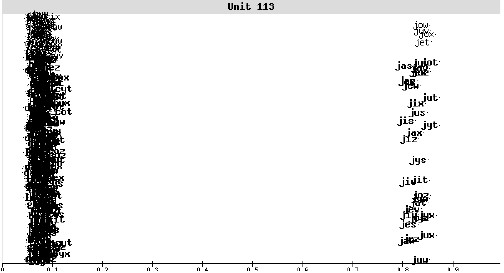}
	\includegraphics[width=132px]{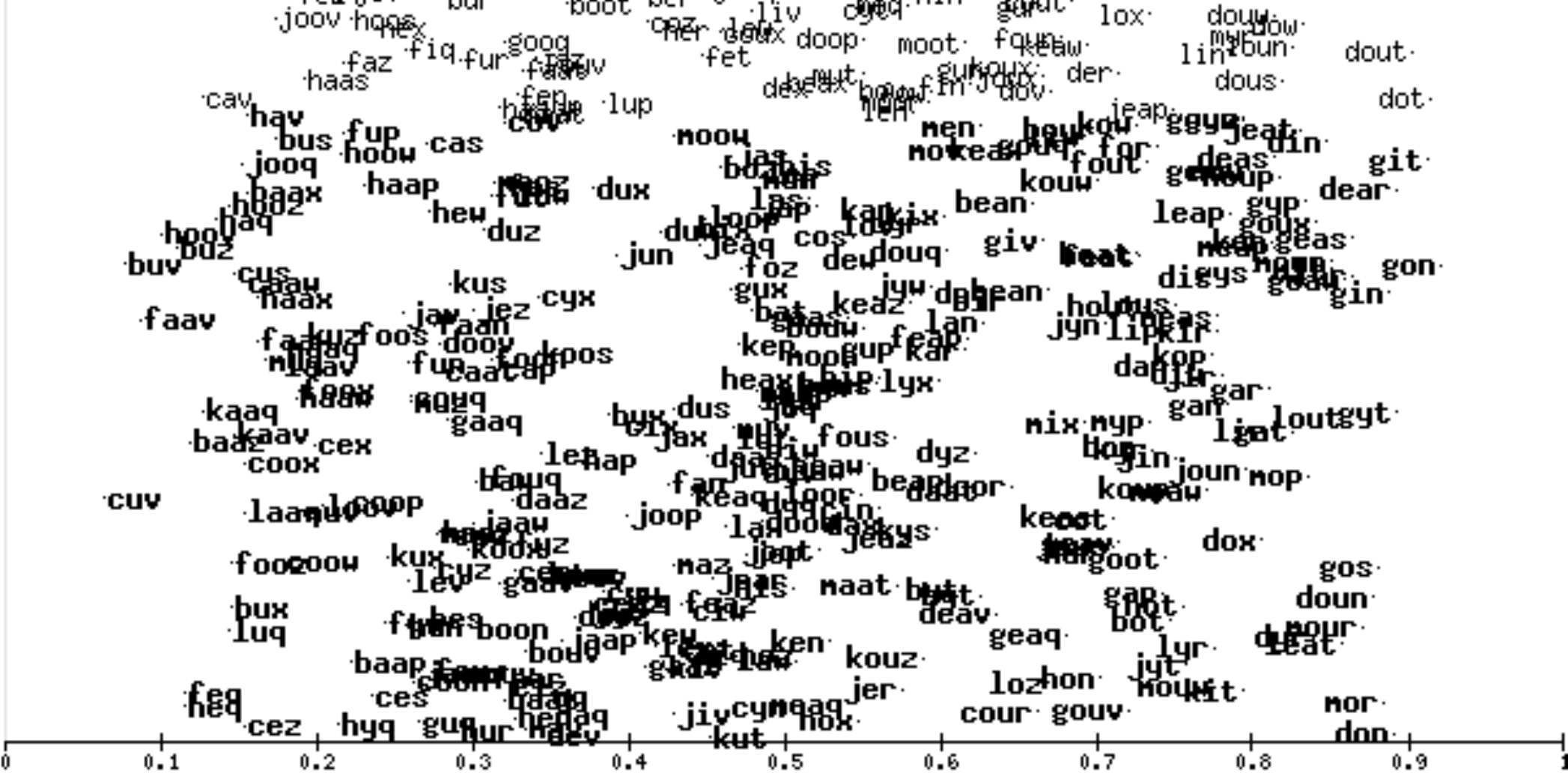}
	\includegraphics[width=68px]{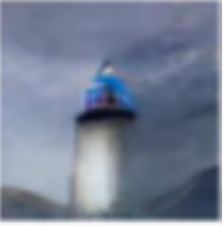}\\
	\includegraphics[width=340px]{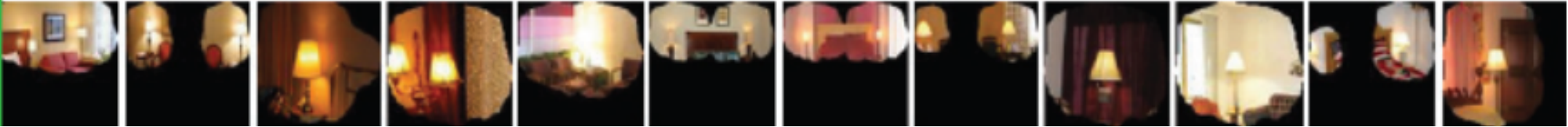}
	\caption{
    \textbf{Top left:} jitterplot of unit \unit{113} in an RNN (under the superposition constraint) selective to the letter `j' \citep{6}. 
    \textbf{Top centre:} jitterplot of a non-selective unit \unit{160} found in an RNN trained on words one-at-a-time from \citep{6}. 
    \textbf{Top right:} Activation maximization image of unit conv5.9 AlexNet that resembles a lighthouse \citep{28}. 
    \textbf{Bottom:} highest-activation images for a `lamp' detector with .84 \precision in the layer \layer{conv5} of AlexNet; from \citep{105}.
	}
	\label{fig:BowersEg}
\end{figure}


Here we compare a range of measures of selectivity on a number of different convolutional networks of object identification, but focus on AlexNet \citep{106} trained on the ImageNet dataset \citep{ImageNet} because many authors have studied the selectivity of single hidden units in this model using a range of quantitative \citep{102, 105} and qualitative \citep{40, 103, 104} methods.  AlexNet is one of the first modern DCNNs, and its dramatic success in categorizing images from ImageNet (that includes over 1 million images of objects and animals taken from 1000 categories) is often credited with starting the modern era of NN research. Its architecture is given in Figure~\ref{fig:AlexNetArch}.  The network includes alternating convolutional and pooling layers from the input up to 5$^{\mathrm{th}}$ convolutional or `conv5' layer. The convolutions are spatially organized learned filters (single units) that encode features within their receptive field, with each filter repeated across multiple spatial locations (analogous to a simple cell in V1 that encodes for a feature in its receptive field -- something like a line of a specific orientation -- with equivalent simple cells coding for the same orientation repeated over retinal locations). Following most of the convolutional layers in AlexNet is a pooling layer, in this case, max pooling, in which units take on the maximum activation value of a given convolutional filter in its receptive field (much like a complex cell in V1 that responds to the most active simple cell within a small retinotopic range). Together the convolutional and pooling layers learn useful visual features for object categorization.  These features are input to layer \layer{fc6} that is the first of three fully connected layers (fc6, \layer{fc7} , \layer{fc8} ) with \layer{fc8} , after applying softmax, encoding all 1000 categories in a localist or `one hot' coding manner (i.e. the `tiger sharks' category is encoded by: [0,0,0,1,0...0]).    

In the experiments reported below, we explored the selectivity for the learned object categories in the last three hidden layers of AlexNet, namely \layer{conv5} , \layer{fc6} and \layer{fc7} . We also assessed the selectivity of units in two more recent DCNNs, namely, VGG-16 and GoogLeNet models trained on the ImageNet and Places-365 dataset (over two million images that depict different scenes; e.g., kitchen, bedroom, forest, etc.).  In these cases, we only consider a few units that were considered  highly selective according to the Network Dissection method \citep{102}. 

In order to directly compare and have a better understanding of the different selectivity measures, we assessed (1) localist, (2) \precision, and (3) CCMAS selectivity, as well as a range of signal detection methods, namely, (4) recall with 100\% and 95\% precision, (5) maximum informedness, (6) specificity at maximum informedness, and (7) recall (also called \emph{sensitivity}) at maximum informedness, and false alarm rates at maximum informedness (all described in Sec.~\ref{sec:methods}). 
In addition to these quantitative measures, we assessed the human interpretation of images generated by a state-of-the-art activation maximization (AM) method \citep{nguyen2017plug} for units in layers \layer{conv5} , fc6, and \layer{fc8}  layers as well as display jitterplots of some of the most selective units as determined by quantitative methods above. 
The jitterplots provide a more intuitive assessment of degree of selectivity that are usefully compared to the different quantitative and AM measures in order to get a better sense of these measures.  

\section{Methods}
\label{sec:methods}

\textbf{Network and Dataset} All $\sim$1.3M photos from the ImageNet ILSVRC 2012 dataset \citep{ImageNet} were cropped to $277\times 277$ pixels and classified by the pre-trained AlexNet CNN \citep{106} shipped with Caffe \citep{jia2014caffe}, resulting in 721,536 correctly classified images. Once classified, the images are not re-cropped nor subject to any changes. To get the activations we fed the correct images into AlexNet and recorded the activations at that layer (futher details and full codebase available at: \url{https://github.com/ellagale/testing_object_detectors_in_deepCNNs}). We analyzed the fully connected (\layer{fc}) layers: \layer{fc6} and \layer{fc7} (4096 units each), and the top convolutional layer \layer{conv5} which has 256 filters. We only recorded the activations of correctly classified images. The activation files are stored in .h5 format and are available at  \url{https://bristol.codersoffortune.net/AlexNet\_Merged/}. 
We randomly selected 233 \layer{conv5}, 2738 \layer{fc6}, 2239 \layer{fc7} units for analysis, amounting to around 90\% of \layer{conv5}, and roughly 60\% of \layer{fc6} and \layer{fc7}, numbers chosen owing to time constraints.


\textbf{Localist selectivity} Following \citet{1}, we define a unit to be localist for class $A$ if the set of activations for class $A$ was higher and disjoint with those of $\neg A$.
Localist selectivity is easily depicted with jitterplots \citep{10} in which a scatter plot for each unit is generated (see Figs.~\ref{fig:BowersEg} and ~\ref{fig:butterfly}). Each point in a plot corresponds to a unit's activation in response to a single image, and only correctly classified images are plotted. The level of activation is coded along the $x$-axis, and an arbitrary value is assigned to each point on the $y$-axis.

\textbf{Precision} \precision refers to the proportion of items above some threshold from a given class. The \precision method of finding object detectors involves identifying a small subset of images that most strongly activate a unit and then identifying the critical part of these images that are responsible for driving the unit. \citet{105} took the 60 images that activated a unit the most strongly and asked independent raters to interpret the critical image patches (e.g., if 50 of the 60 images were labeled as `lamp', the unit would have a\precision index of 50/60 or .83; see Fig.~\ref{fig:BowersEg}). Object detectors were defined as units with a \precision score $>$ .75: they reported multiple such detectors. Here, we approximate this approach by considering the 60 images that most strongly activate a given unit and assess the highest percentage of images from a given output class.


\textbf{CCMAS} \citet{101} used a selectivity index called the Class-Conditional Mean Activation Selectivity (CCMAS). The CCMAS for class $A$ compares the mean activation of all images in class $A$, $\mu_A$, with the mean activation of all images not in class $A$, $\mu_{\neg A}$, and is given by: $\left( \mu_A - \mu_{\neg A} \right) / \left( \mu_A + \mu_{\neg A} \right)$. 
Here, we assessed class selectivity for the highest mean activation class. 

\textbf{Activation Maximization} We harnessed an activation maximization method called Plug \& Play Generative Networks (PPGNs) \citep{nguyen2017plug} in which an image generator network was used to generate images (AM images) that highly activate a unit in a target network. 

Formally, we attempt to maximize the activation $\phi(.)$ of a neuron indexed $k$ at layer $l$ of a target neural network:

\begin{equation}
    x^* = \mathrm{arg max}_{x} (\phi_{l, k}(x))
\end{equation}

However, simply modifying an image pixel-wise in the direction of increasing neural activity often yields similar and noisy stimuli that are not human-interpretable \cite{nguyen2019understanding}.
Therefore, PPGNs authors proposed to harness an image generator network $G$ as a strong natural image prior and search in the input space of generator $G$ for input vectors $z$ is in $R^{4096}$ such that the generated images $G(z)$ do not only (1) cause high neural activation but are also (2) realistic and (3) diverse \cite{nguyen2017plug}. We used the public PPGN code released by \citet{nguyen2017plug} and their default hyperparameters.\footnote{\url{https://github.com/Evolving-AI-Lab/ppgn}} That is, we generated each image by running an Stochastic Gradient Descent (SGD) optimizer for 200 steps with an initial learning rate of 1.0, and the multipliers for the realism, high-activation, and diversity terms are $10^{-5}$, 1, and $10^{-17}$, respectively. We generated 100 separate images that maximally activated each unit in the \layer{conv5}, \layer{fc6}, and \layer{fc8} layers of AlexNet. 
Images were used in the experiment described below (Sec.~\ref{sec:AM}). 


\textbf{Recall with perfect and 95\% precision} Recall with perfect and 95\%\precision are related to localist selectivity except that they provide a continuous rather than discrete measure.  For recall with perfect\precision we identified the image that activated a given unit the most and counted the number of images from the same class that were more active than all images from all other classes.  We then divided this result by the total number of correctly identified images from this class.  A recall with a perfect\precision score of 1 is equivalent to a localist representation.  Recall with a 95\%\precision allows 5\% false alarms.

\textbf{Maximum informedness}  Maximum informedness identifies the class and threshold where the highest proportion of images above the threshold and the lowest proportion of images below the threshold are from that class \citep{powers2011evaluation}.  The informedness is computed for 
each class at each threshold, with the highest value selected. Informedness summarises the diagnostic performance of unit for a given class at a certain threshold based on the recall 
$[\mathrm{True Positives} / (\mathrm{True Positives} + \mathrm{False Negatives})]$
and specificity $[\mathrm{True Negatives} / \mathrm{True Negatives} + \mathrm{False Positives})]$ in the formula $[\mathrm{informedness} = \mathrm{recall} + \mathrm{specificity} - 1]$ \citep{powers2011evaluation}. 

\textbf{Sensitivity or Recall at Maximum Informedness} For the threshold and class selected by Maximum Informedness, recall (or hit-rate) is the proportion of items from the given class that are above the threshold. Also known as true postive rate.

\textbf{Specificity at Maximum Informedness} For the threshold and class selected by Maximum Informedness, the proportion of items that are not from the given class that are below the threshold. Also known as true negative rate.


\textbf{False Alarm Rate at Maximum Informedness} For the threshold and class selected by Maximum Informedness, the proportion of items that are not from the given class that are above the threshold.

\textbf{Network Dissection} 
Network Dissection \cite{bau2017network} is a method for assessing the selectivity of convolutional filter with respect to over a thousand visual concepts relating to scenes, objects, parts, materials, colours and textures as coded in the Broden dataset  \citep{102}.  The Broden dataset contains 60000 real-world images, each with an accompanying concept-location map, coding at the pixel-level where a given concept occurs in the image.  For example, if the concept is the colour `red', all red pixels in the image will be labelled 1 on the concept-location map and all other pixels with be 0.  Network Dissection compares the concept-location map for an image with the activation map of a convolutional filter in response to that image. 
This comparison is done using intersection over union (IoU):

\begin{math}
IoU = \frac{\mathrm{the number of pixel locations that are 1 in both maps}}{\mathrm{the total number of unique pixels labelled 1 in both maps}}
\end{math}

If the IoU score is greater that .04, then the filter that produced the activation map is labelled as a detector for the labelled concept. Note, we did not carry out any network dissection analyses ourselves, but simply selected units that were considered object detectors according to this metric by \citet{102} in Section 3.3 and \ref{sec:test_other_models}.



\textbf{Methodological details for the behavioral experiment\label{sec:psychometh}} One hundred generated images were made for units in hidden layers \layer{conv5} and \layer{fc6} and output layer \layer{fc8} in AlexNet, as in Nguyen et al. (2017), and displayed as $10\times10$ image panels. 
We chose these three layers because they span across a wide spectrum of neural selectivity and two main types of layers: convolutional and fully-connected.
\layer{conv5} were found to contain high-level object detectors (e.g., dog faces) in convolutional layers \cite{bau2017network,zhou2014learning}.
\layer{fc6} is a fully-connected layer that contain units often capture amalgamation of different concepts (i.e., a generalist rather localist neurons) \cite{nguyen2016multifaceted}.
\layer{fc8} neurons are trained specifically to light up for images of pre-defined categories and therefore are expected to exhibit a high degree of selectivity.

A total of 3,299 image panels were used in the experiment (995 associated with \layer{fc8} output units with 5 units omitted by mistake, all 256 \layer{conv5} units, and 2048 randomly selected \layer{fc6} image panels constituting half of all units in this layer) and were divided into 64 counterbalanced lists of 51 or 52 (4 \layer{conv5}, 15 or 16 \layer{fc8} and 32 \layer{fc6}). Fifty-one of the lists were assigned to 5 participants and 13 lists were assigned to 6 participants. The study was approved by the University of Bristol Faculty of Science Ethics Committee and informed consent was obtained from all participants. 

To test the interpretability of these units, paid volunteers were asked to look at image panels and asked if the images had an object / animal or place in common, i.e. a concrete object.  In training, they were also shown examples of panels that only included common abstract concepts, like `color', `shape' or `texture', that required a `no' for an answer. If the answer was `yes', they were asked to name that object simply (i.e. fish rather than goldfish). For any units where over 80\% of humans agreed that there was an object present, analyses of the common responses were carried out by reading the human responses and comparing them to both each other and to the output class labels. Agreement was taken if the object was the same rough class. For example, `beer', `glass', and `drink' were all considered to be in agreement in the general object of `drink', and in agreement with both the classes of `wine glass' and `beer' as these classes were also general drink classes (this is an actual example, most responses were more obvious and required far less interpretation than that). Participants were given six practice trials, each with panels of 20 images before starting the main experiment. Practice trials included images that varied in their interpretability.  Analyses of common responses were done for any units where over 80\% of humans agreed there was an object present. An illustration of the task can be found in Appendix A, and readers can test themselves at:
\url{https://research.sc/participant/login/dynamic/63907FB2-3CB9-45A9-B4AC-EFFD4C4A95D5}. All materials used in the AM experiment are stored here: \url{https://gorilla.sc/openmaterials/84689}.

\section{Results}

\subsection{Comparison of selectivity measures in AlexNet}
\label{sec:PcE}

\begin{figure*}[htb]
\centering
\begin{tabular}{ccc}
    \centering
    a. Precision & b. No. of classes in top 100 & c. CCMAS \\
    \includegraphics[width=0.29\linewidth]{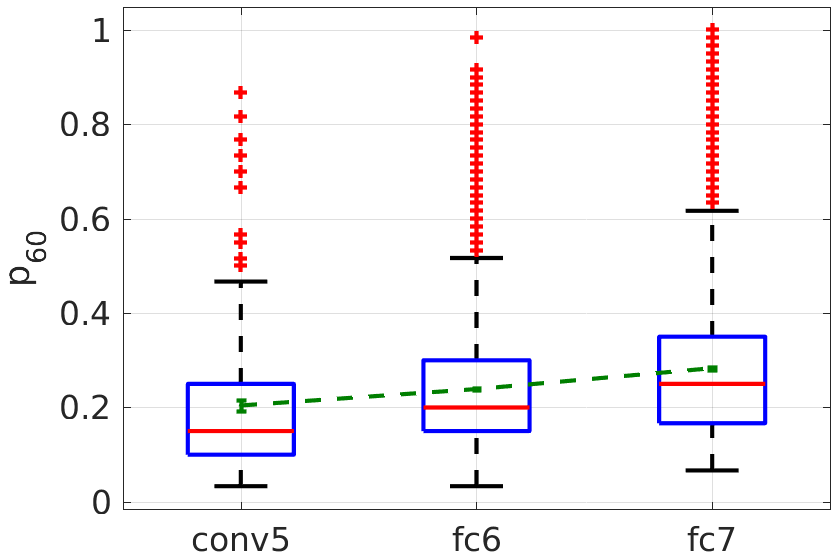} & 
    \includegraphics[width=0.29\linewidth]{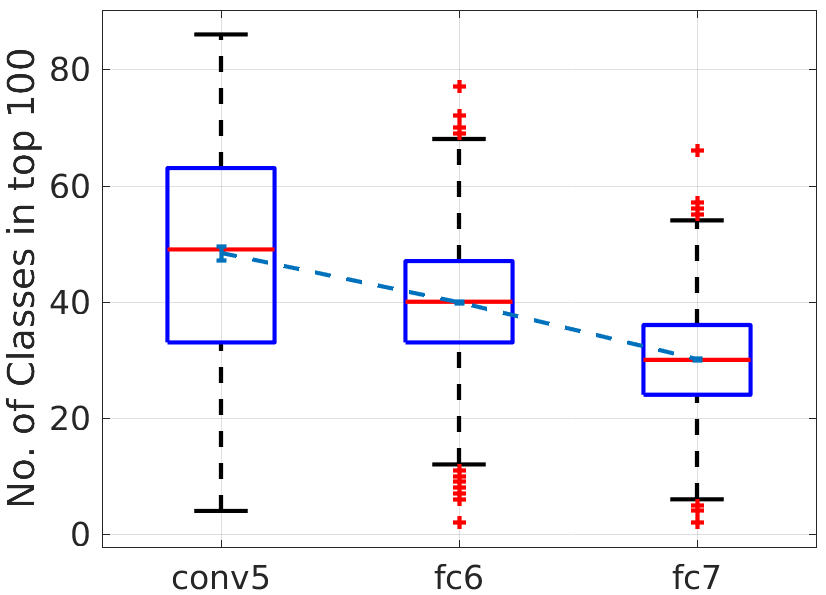} &
    \includegraphics[width=0.29\linewidth]{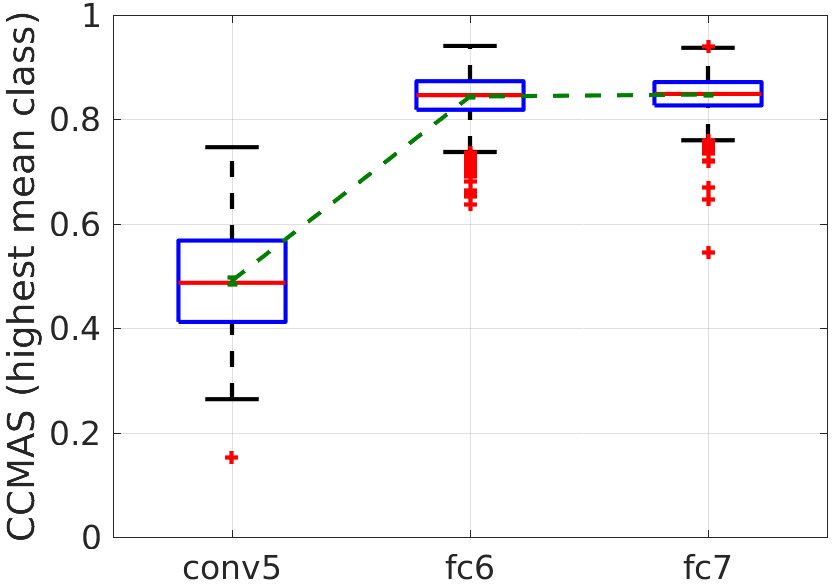} \\
    d. Recall (precision = 1) & e. Recall (precision = 0.95) & f. Max. informedness  \\
    \includegraphics[width=0.29\linewidth]{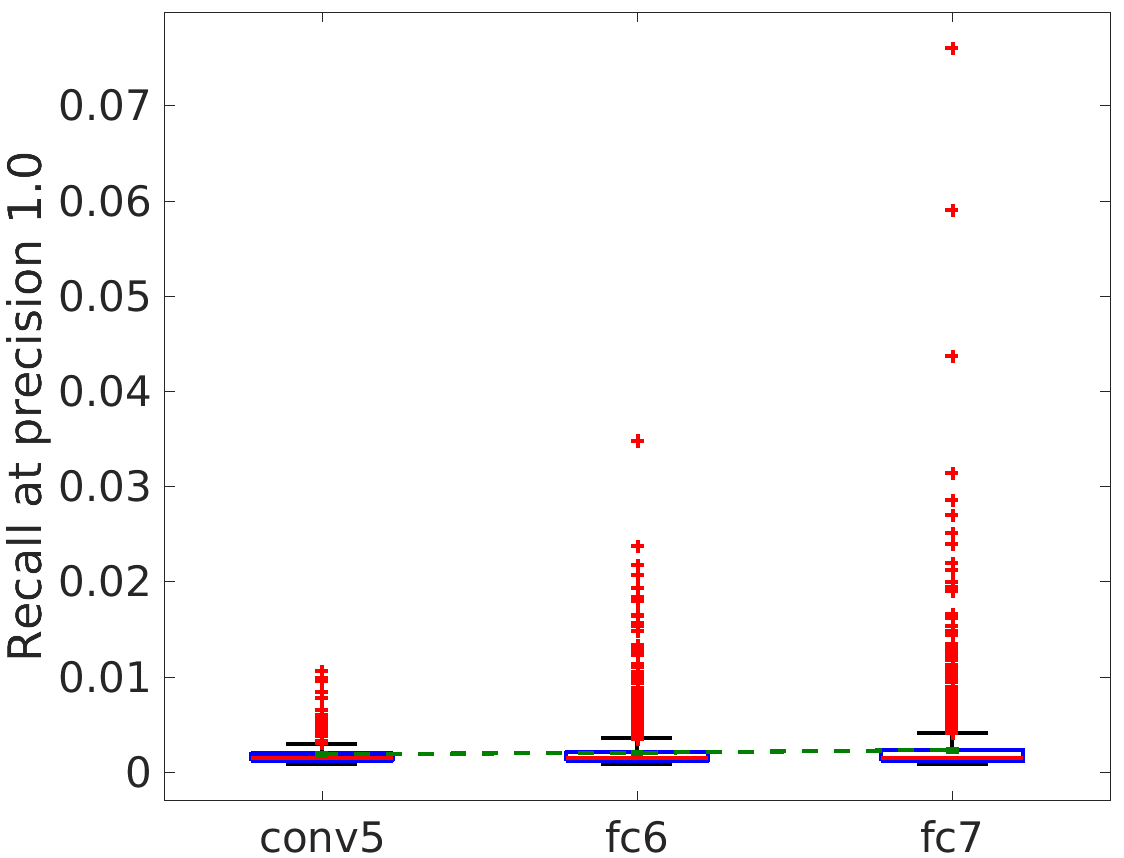} &
	\includegraphics[width=0.29\linewidth]{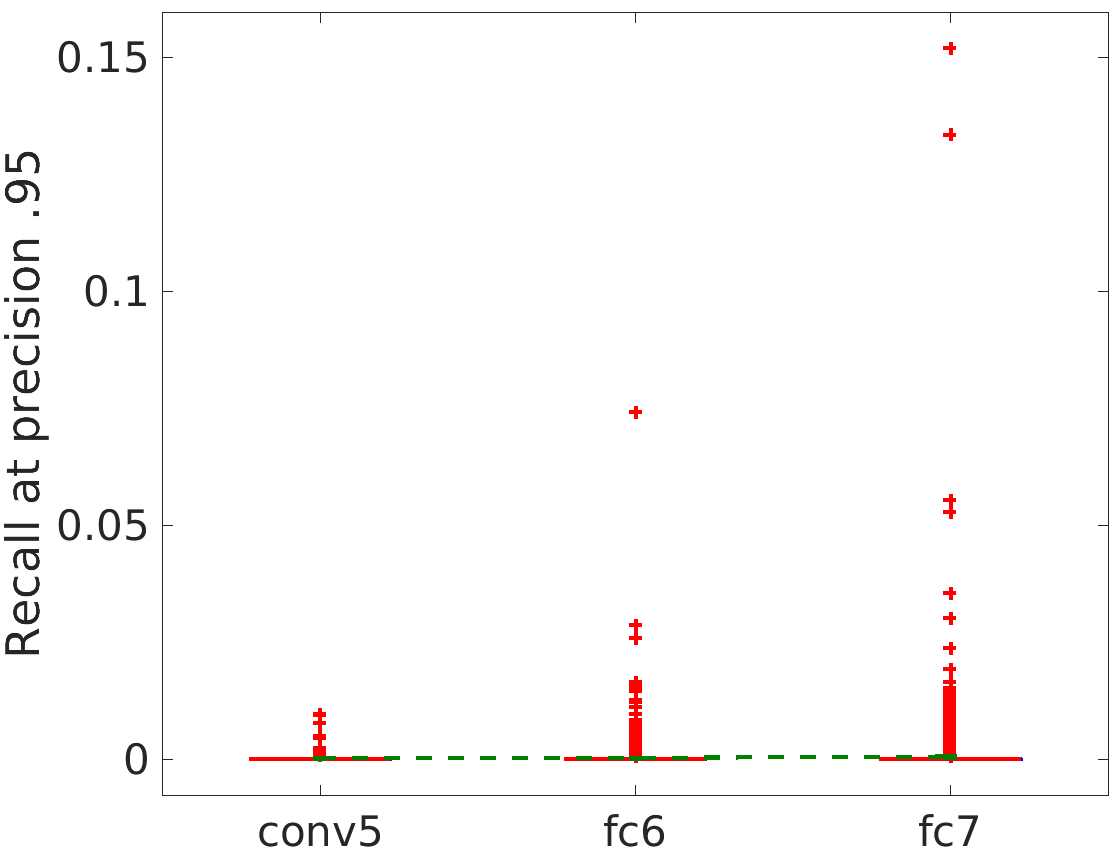} &
	\includegraphics[width=0.29\linewidth]{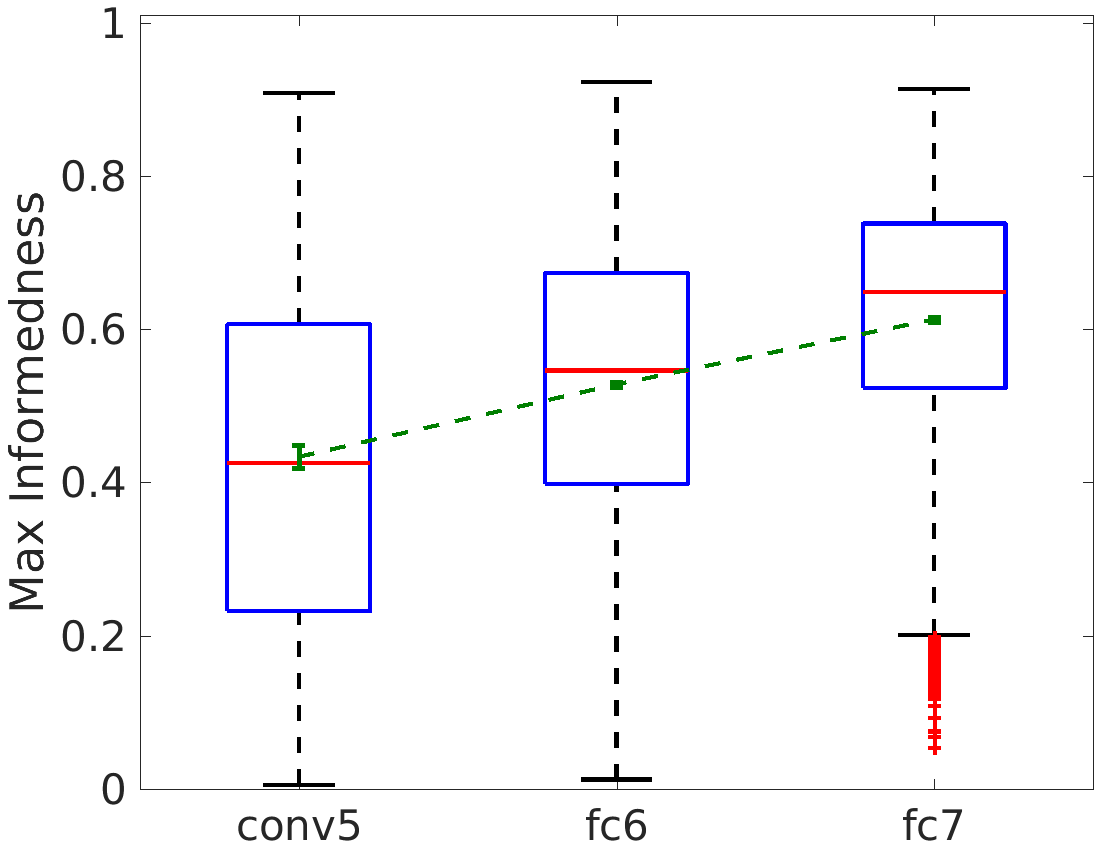}  \\
	g. Specificity & h. Recall  & i. False alarm proportion \\
	(at max. informedness) &  (at max. informedness)  & (at max. informedness)\\ 
	\includegraphics[width=0.29\linewidth]{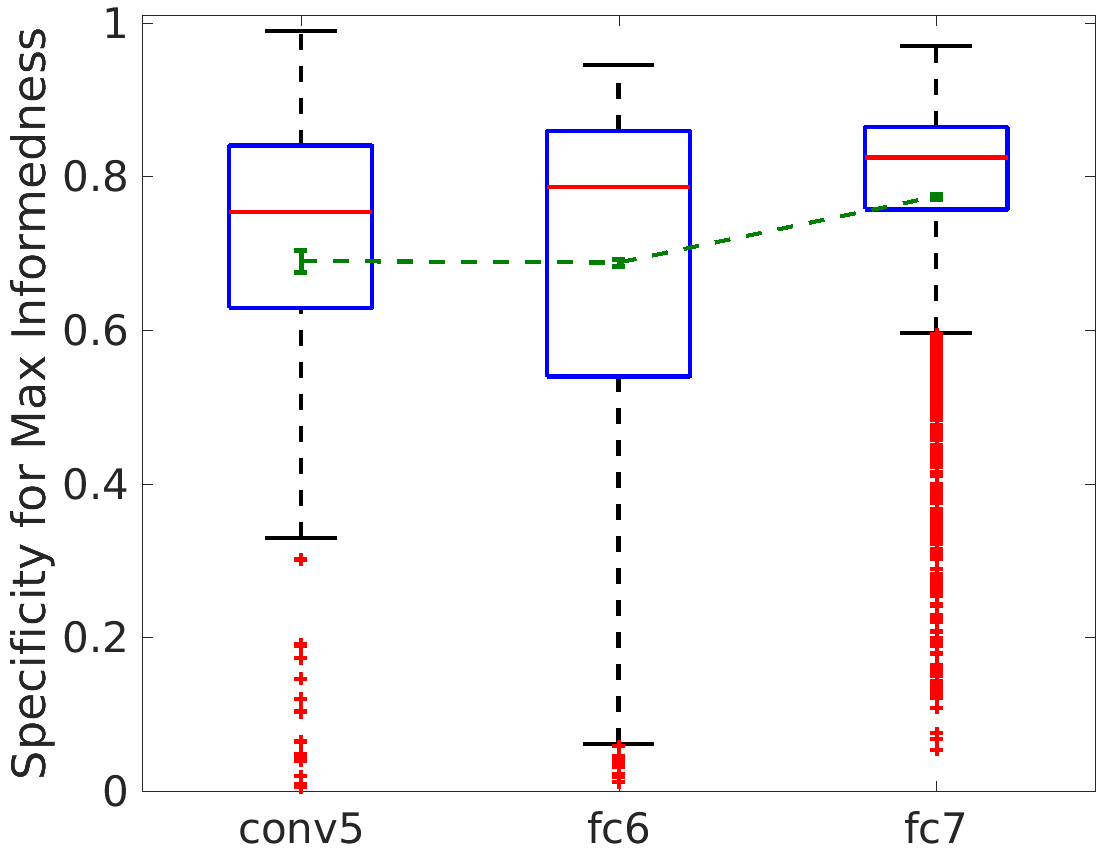} &
	\includegraphics[width=0.29\linewidth]{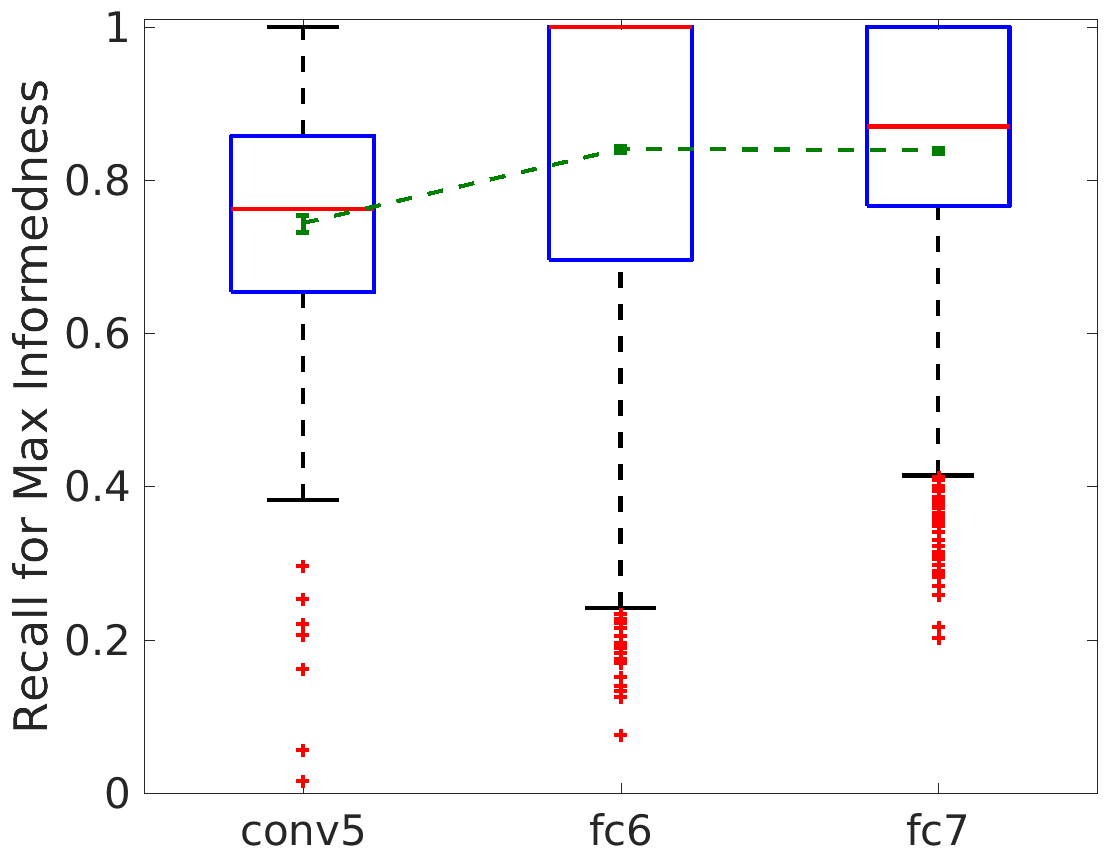} &
	\includegraphics[width=0.29\linewidth]{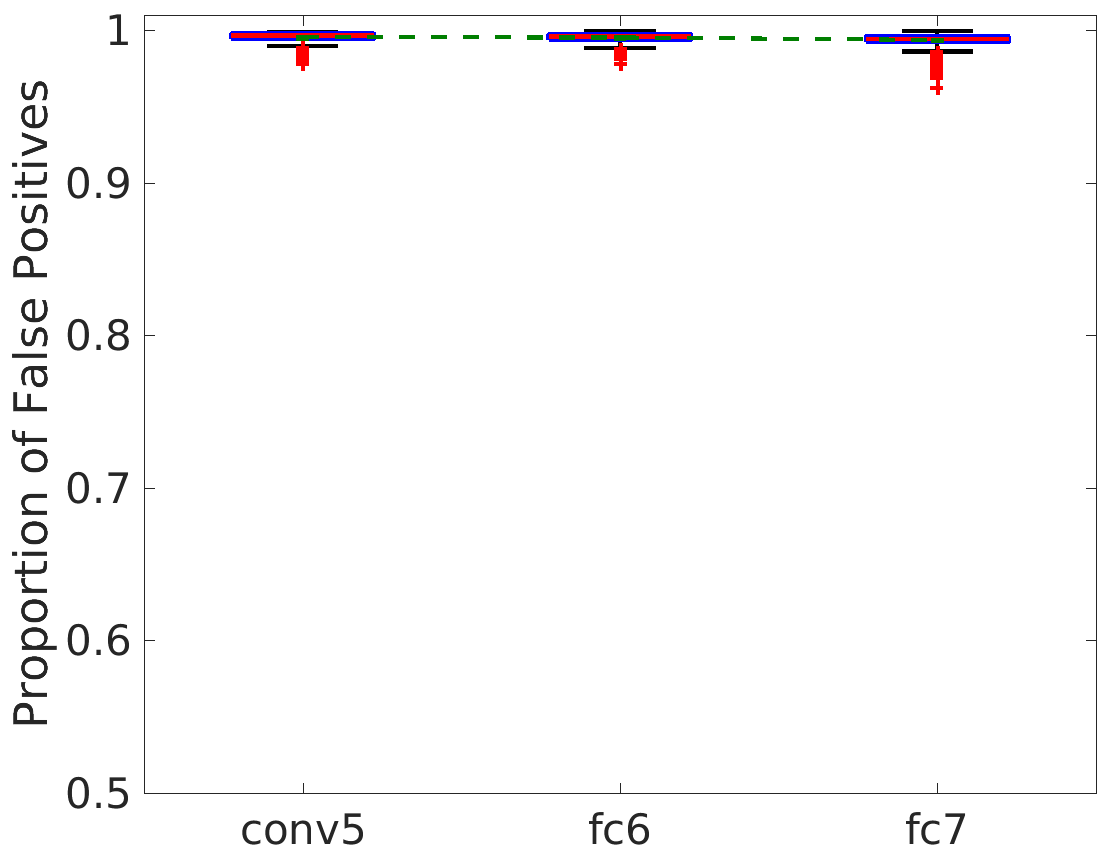} \\

\end{tabular}
\caption{Different selectivity measures across the \layer{conv5}, \layer{fc6}, and \layer{fc7} layers of AlexNet. Red-line: median of data, top and bottom of box edges is the 25$^{\mathrm{th}}$ and 75$^{\mathrm{th}}$ percentile, whiskers extend to extreme edges of distribution not considered outliers and red crosses are outliers. Green points and dashed lines are the means of the distributions with standard errors. The high levels of selectivity observed with the \precision and CCMAS measures are in stark contrast with the low levels of selectivity observed with the recall with perfect\precision and high false-alarm rates at maximum informedness. Note the y-axis scaling for panels 3e, f, and i are different from other panels in order to depict the findings more clearly.    
}
\label{fig:allthedata}
\end{figure*}

\begin{figure*}[htb]
    \centering
	\includegraphics[width=0.33\linewidth]{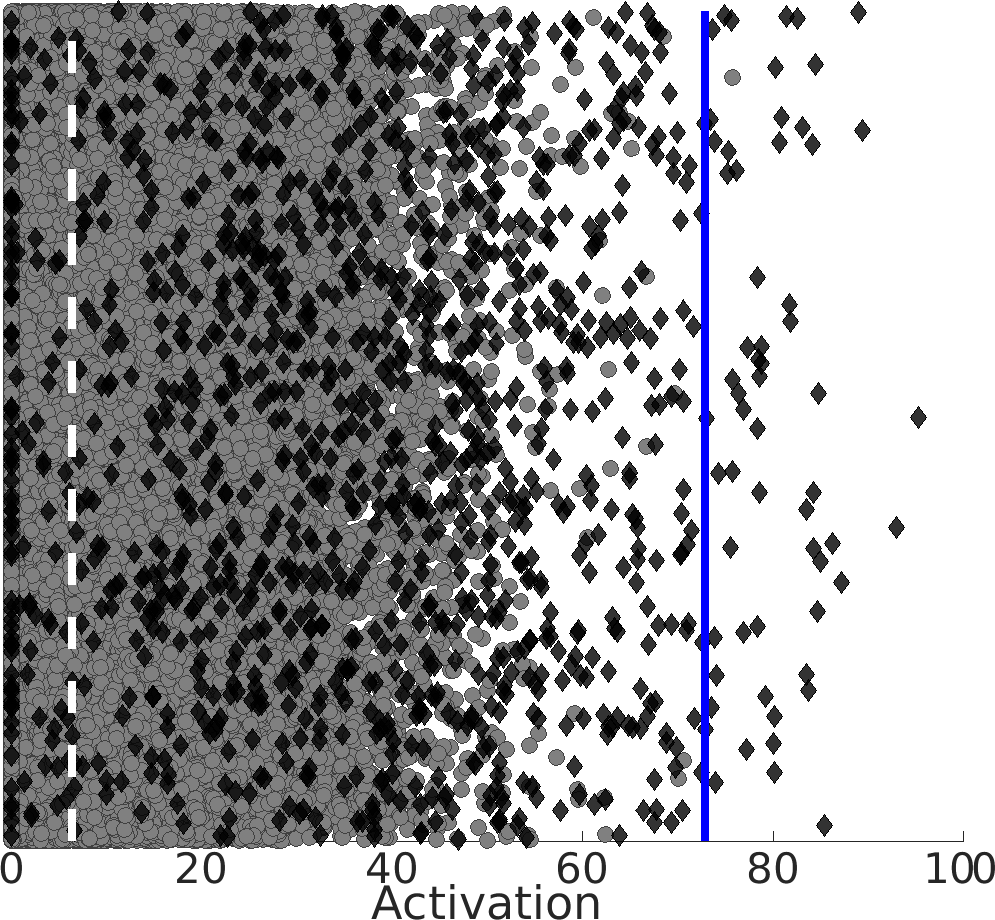}
		\includegraphics[width=0.34\linewidth]{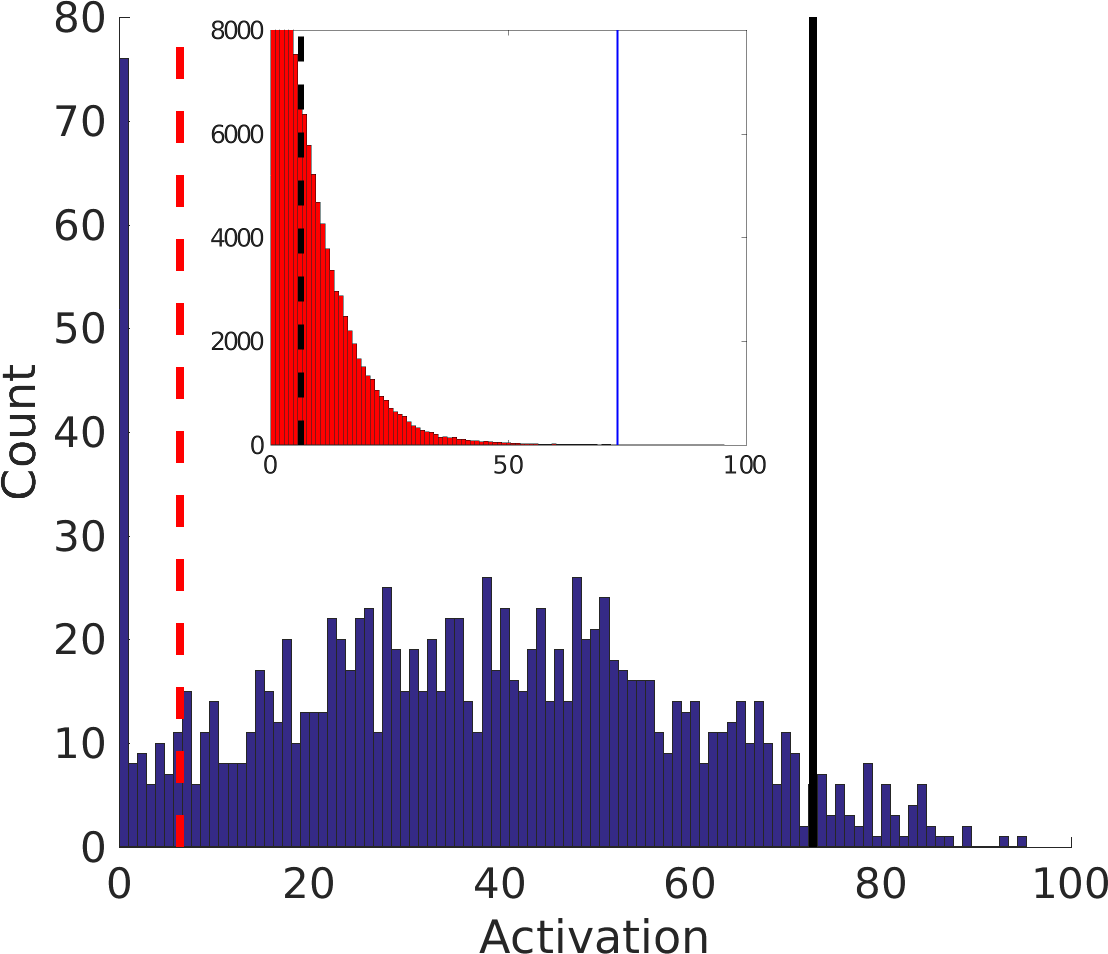}
		\includegraphics[width=0.27\linewidth]{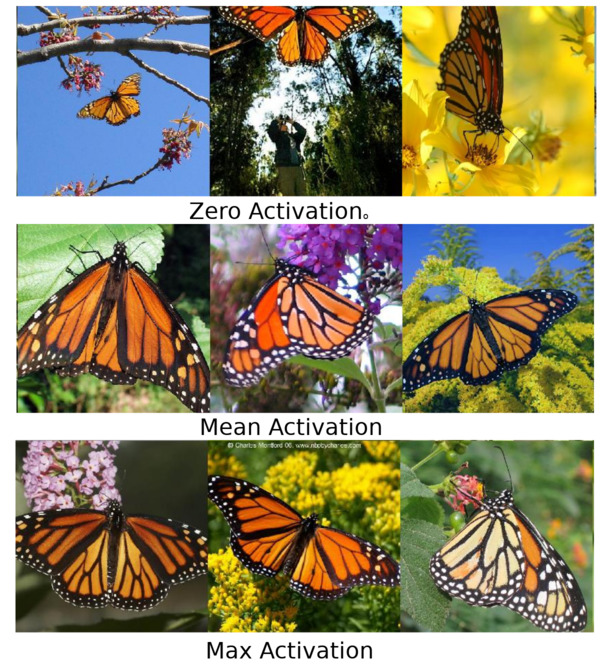}
    \caption{Data for unit fc6.1199. \textbf{Left:} activation jitterplot, black diamonds: Monarch butterfly images; grey circles: all other classes; white dashed line: threshold for the butterfly class maximum informedness; blue solid line: threshold for top 60 activations. \textbf{Middle:} histogram of activations of Monarch butterflies; red dashed line: threshold for the butterfly class maximum informedness; black solid line: threshold for top 60 activations.  \textbf{Inset:} zoomed-in histogram of all activations across all ImageNet classes of unit fc6.1199 (N.B. this plot shows only the highest 121,586 activations; there are 596,734 activations at 0). There are Monarch butterfly images covering the whole range of values, with 72 images (5.8\% of the total) having an activation of 0. \textbf{Right:} example ImageNet images with activations of 0 (top), the mean, 39.2$\pm$0.6, (middle), and the maximum, 95, (bottom) of the range. Although the high \precision score suggests that this unit is a butterfly detector this is misleading given there are butterfly images over the entire activation range (including 0).
}
\label{fig:butterfly}
\end{figure*}

The results from the various selectivity measures applied to the \layer{conv5}, \layer{fc6}, and \layer{fc7} layers of AlexNet are displayed in Fig.~\ref{fig:allthedata}a--i. We did not plot the localist selectivity as there were no localist `grandmother units'. The first point to note is that multiple units in the \layer{fc6} and \layer{fc7} layers had \precision and CCMAS scores approaching 1.0. 
For example, in layer \layer{fc7}, we found 14 units with a \precision > 0.9, and 1487 units with a CCMAS > 0.9.  The second point is that other measures highlight much reduced estimates of selectivity.  For example, unit \layer{fc7}.\unit{255} had a CCMAS of .9 and a\precision of .97, but its recall with a perfect \precision score was only .08, meaning that there was at least one non-Monarch butterfly image more strongly activated than 92\% of the Monarch butterfly images (and this was the highest recall with a perfect \precision score in the model). A similar pattern of results was observed with recall with .95 precision, as shown in panel 3e.

The unit with the top maximum informedness score (unit \unit{3290} also responding to images of Monarch butterflies with a score of 0.91) had a false alarm rate above its optimal threshold > 99\% (indeed the minimum false alarm rate for any unit was 0.96). This means that over 99\% of images that activate this unit above its ideal threshold for detecting Monarch butterflies, are \textit{not} Monarch butterflies.

To illustrate the contrasting measures of selectivity consider unit fc6.1199 depicted in Fig.~\ref{fig:butterfly} that has a\precision score of .98 and a CCMAS score of .92. By \citeauthor{105}'s criterion, this is a `Monarch Butterfly' detector (its \precision score is > .75). The Maximum Informedness score was .82, and again > 99\% of images active above this threshold (white dashed line in Fig.~\ref{fig:butterfly}) were false alarms.  A more conservative threshold would reduce the false alarm rate. For example, setting a threshold below the 60 most active items (blue solid line denoting the the \precision measure threshold) has a false alarm rate of .02.  However, only 59 of the 1241 Monarch butterflies are above this threshold (e.g., sensitivity of .05).  Maximum Informedness scores reflect the trade off between false alarms and false negatives, and as such, gives a lower selectivity score to this unit.  



\subsection{Human interpretation of Activation Maximization images for AlexNet units \label{sec:AM}}

Activation Maximization is one of the most commonly used interpretability methods for explaining what a single unit has learned 
in many artificial CNNs and even biological neural networks (see \cite{nguyen2019understanding} for a survey).
Our behavioral experiment provides the first quantitative assessment of AM images and compares AM interpretability to other selectivity measures.

\begin{table*}[thb]
\small
\caption{Human judgements of whether AM images look like familiar objects in layers \layer{conv5}, \layer{fc6}, and \layer{fc8} in AlexNet.  Standard error shown in parenthesis.} 
\label{tab:HumInt}
\begin{center}
\begin{tabular}{llllll}
\multicolumn{1}{c}{\bf layer}  & \multicolumn{1}{c}{\bf \% `yes'} &  \multicolumn{1}{c}{\bf \% units $\geq$ 80\%} & \multicolumn{3}{c}{\bf \% overlap between humans and: } \\
\multicolumn{1}{c}{\bf }  & \multicolumn{1}{c}{\bf responses} &  \multicolumn{1}{c}{\bf `yes' response} & \multicolumn{1}{c}{\bf humans} & \multicolumn{1}{c}{\bf most active}
& \multicolumn{1}{c}{\bf CCMAS}\\
\multicolumn{1}{c}{\bf }  & \multicolumn{1}{c}{(a)} &  \multicolumn{1}{c}{(b)} & \multicolumn{1}{c}{(c)} & \multicolumn{1}{c}{ object (d)} & \multicolumn{1}{c}{class (e)}\\
\hline 
\layer{conv5}        & 21.7	($\pm$1.1)	& 	4.3 ($\pm$ 1.3) & 	89.5 ($\pm$5.7 )&
34.1 ($\pm$14.4) & 	0 \\
\layer{fc6}          & 21.0 ($\pm$0.4)  & 	3.1 ($\pm$ 0.4) & 	80.4 ($\pm$4.1) & 	23.3 ($\pm$5.9) & 	18.9 ($\pm$5.9)  \\
\layer{fc8} (Output)         & 71.2 ($\pm$0.6)  & 	59.3 ($\pm$1.6) & 96.5 ($\pm$0.4) & 
95.4 ($\pm$0.6) & 	94.6 ($\pm$0.7)  \\
\end{tabular}
\end{center}
\end{table*}

\begin{figure*}[htb]
	\centering
	\begin{subfigure}{0.32\linewidth} 
		\centering
		\includegraphics[width=1.0\linewidth]{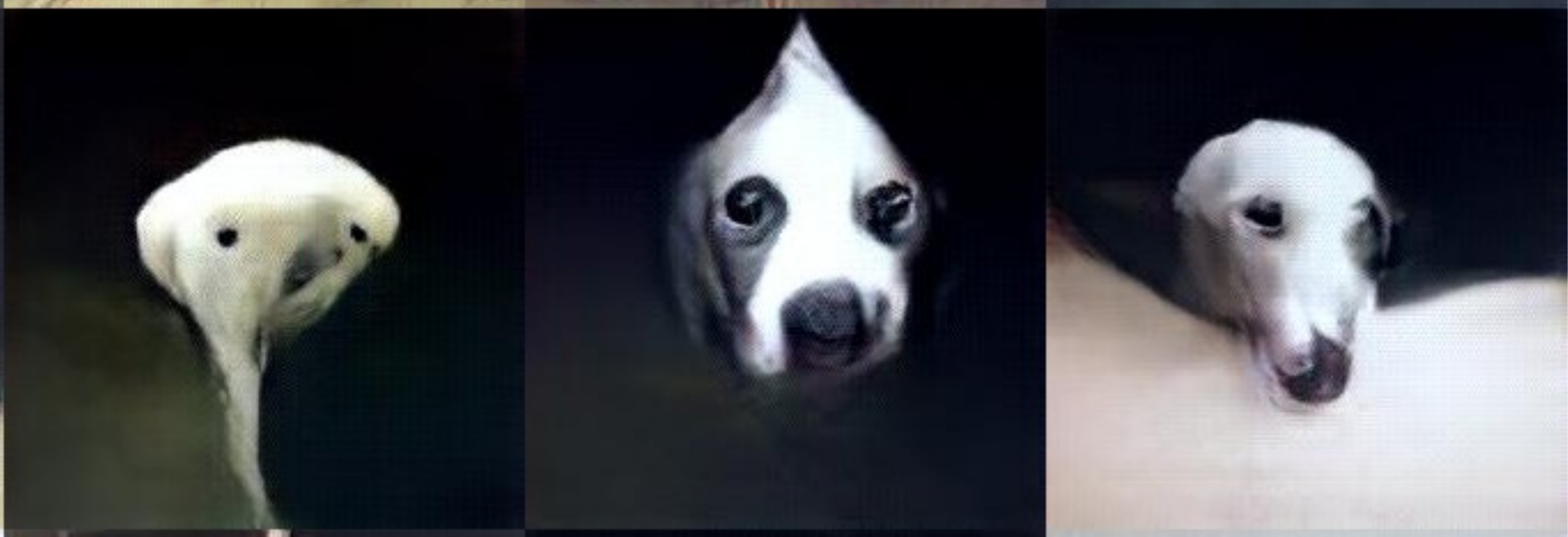}
		\caption{conv5.183}		
	\end{subfigure}	
	\begin{subfigure}{0.32\linewidth} 
		\centering
		\includegraphics[width=1.0\linewidth]{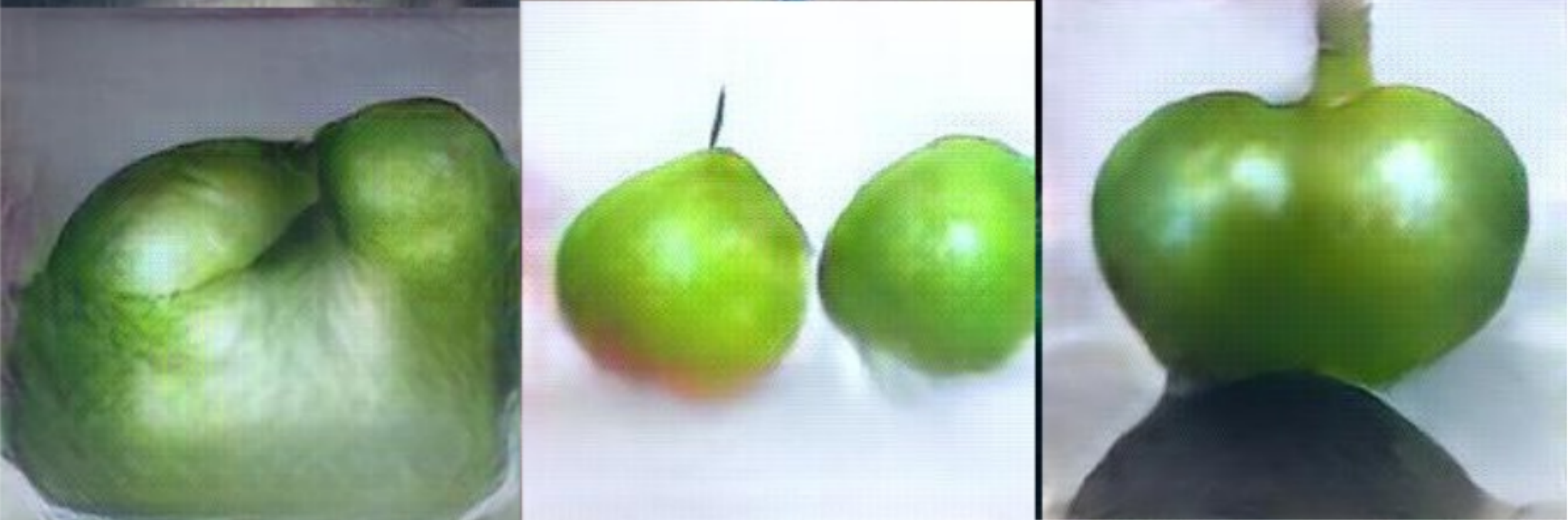}
		\caption{fc6.319}		
	\end{subfigure}	
	\begin{subfigure}{0.32\linewidth} 
		\centering
		\includegraphics[width=1.0\linewidth]{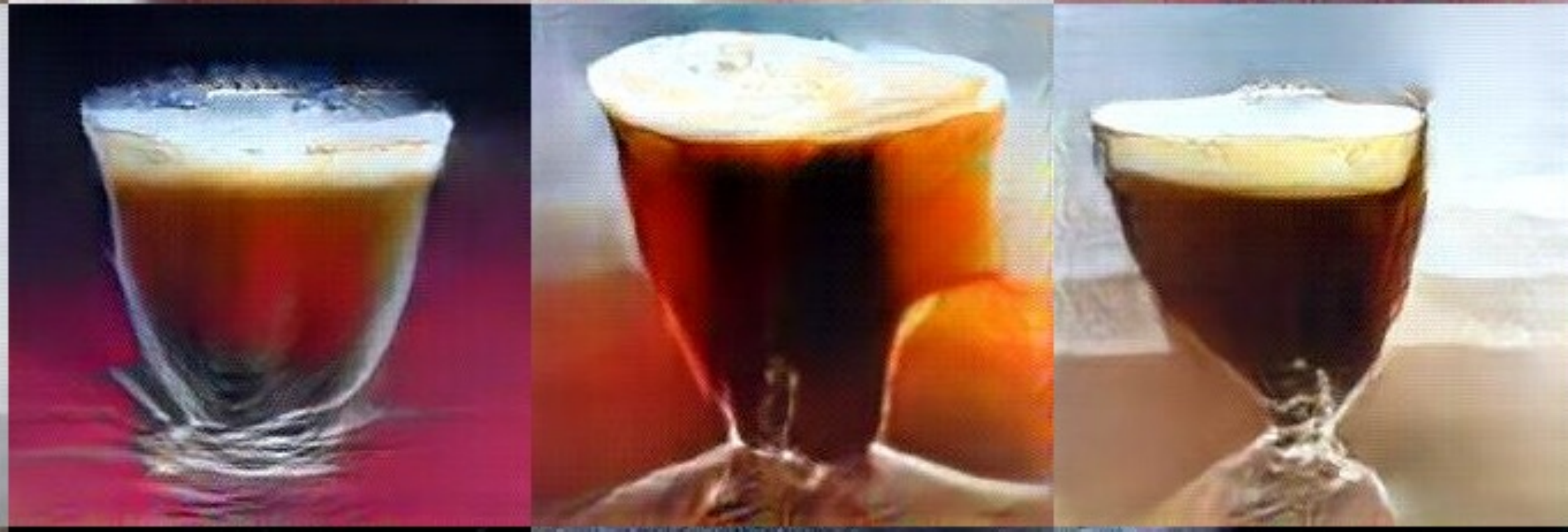}
		\caption{fc8.969}		
	\end{subfigure}	
	\begin{subfigure}{0.32\linewidth} 
		\centering		\includegraphics[width=1.0\linewidth]{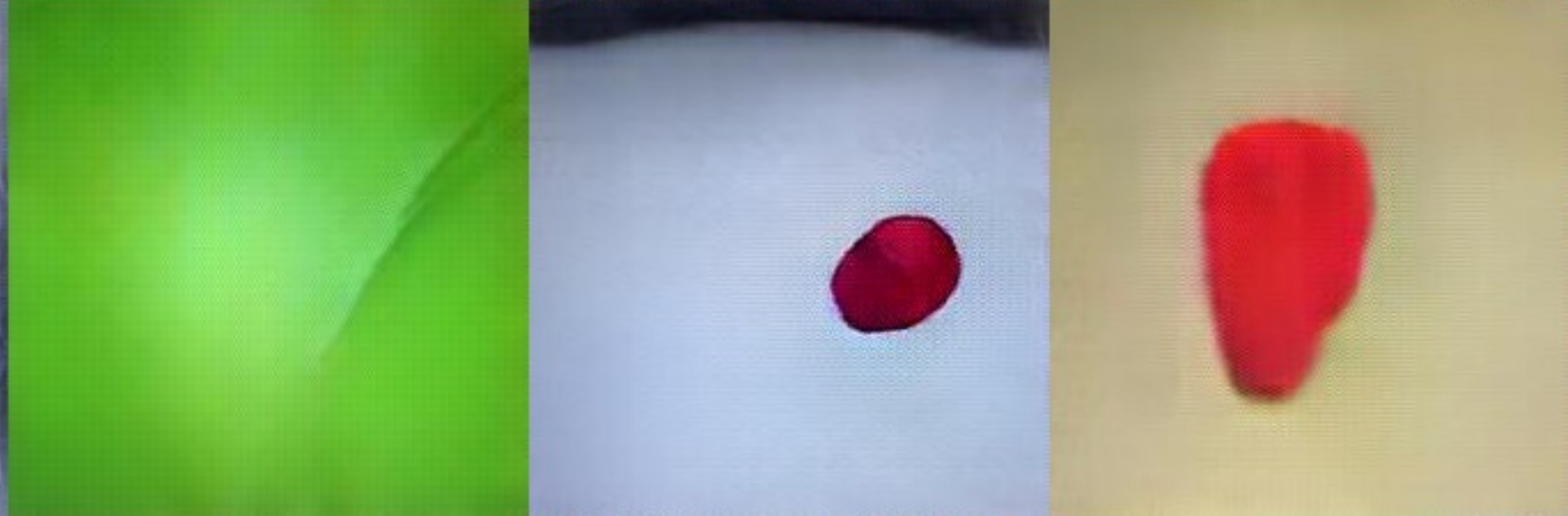}
		\caption{conv5.65}		
	\end{subfigure}	
	\begin{subfigure}{0.32\linewidth} 
		\centering
		\includegraphics[width=1.0\linewidth]{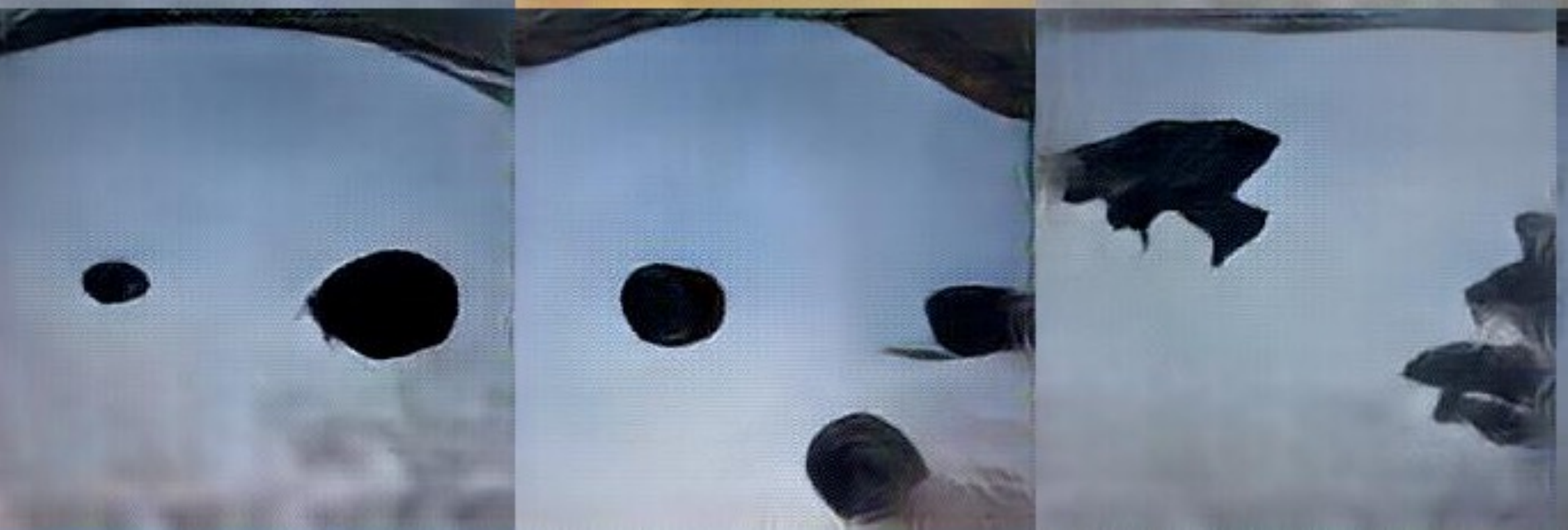}
		\caption{fc6.103}		
	\end{subfigure}	
	\begin{subfigure}{0.32\linewidth} 
		\centering
		\includegraphics[width=1.0\linewidth]{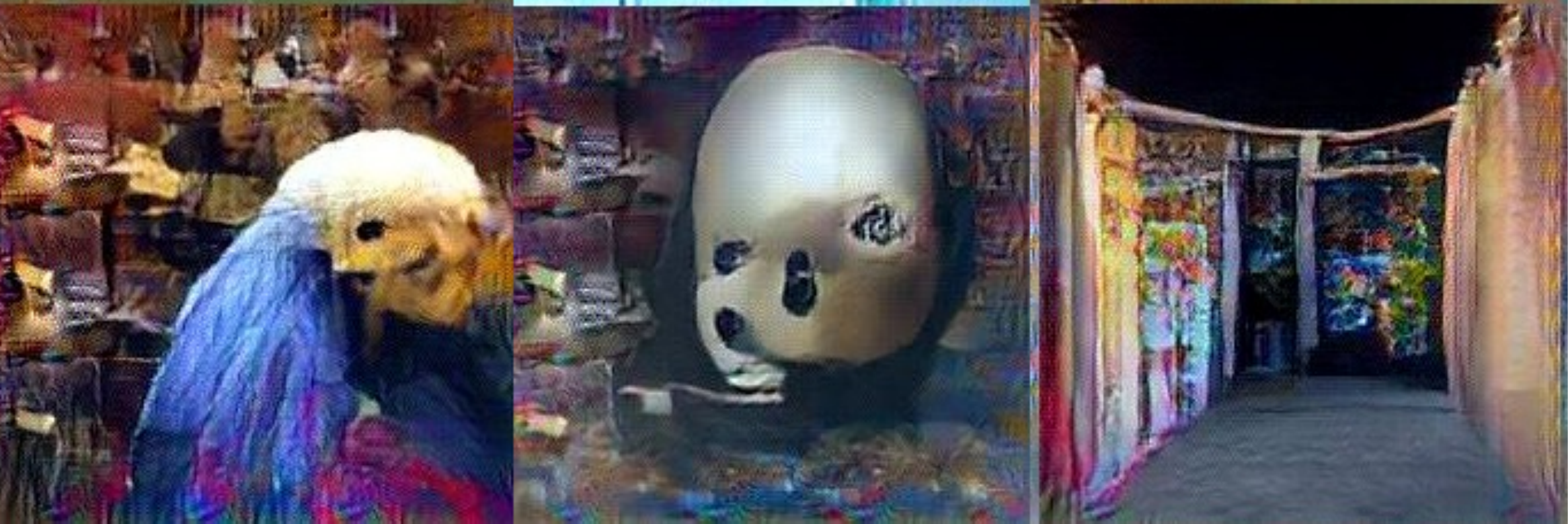}
		\caption{fc8.865}		
	\end{subfigure}	
    \caption{Example AM images that were either judged by all participants to contain objects (a--c) or to be uninterpretable as objects (d--f). The human label for unit conv5.183 (a) was `dogs'; the most active image was of a `flat-coated retriever'; CCMAS class was `monitor'. For fc6.319 (b), subjects reported `green peppers' or `apples' (all classified as the same broad class in our analysis); both the most active item and CCMAS class were `Granny Smith apples'. For fc8.969 (c), humans suggested `beverage' or `drink'; both the most active item and CCMAS class were `eggnog'. 
}
\label{fig:AMObjects}
\end{figure*}


The results are summarized in Table~\ref{tab:HumInt}. 
Not surprisingly, the AM images for output \layer{fc8} units are the most human-recognizable as objects across the AlexNet layers (71.2\%; Table~\ref{tab:HumInt}a).
In addition, when they were given a consistent interpretation, they almost always (95.4\%; Table~\ref{tab:HumInt}d) match the corresponding ImageNet category.  
By contrast, less than 5\% of units in \layer{conv5} or \layer{fc6} were associated with consistently interpretable images (Table~\ref{tab:HumInt}b), and 
the interpretations only weakly matched the category associated with the highest-activation image or CCMAS selectivity (Table~\ref{tab:HumInt}d--e). Apart from showing that there are few interpretable units in the hidden layers of AlexNet, our findings show that the interpretability of images does not imply a high level of selectivity given the signal-detection results (Fig. 2d--h).  
See Fig.~\ref{fig:AMObjects} for an example of the types of images that participants rated as objects or non-objects.


\subsection{Comparing selectivity measures in other CNNs \label{sec:comp}}

\begin{figure*}[hbt]
\centering
\begin{tabular}{ccc}
    \centering
    \includegraphics[width=0.31\linewidth]{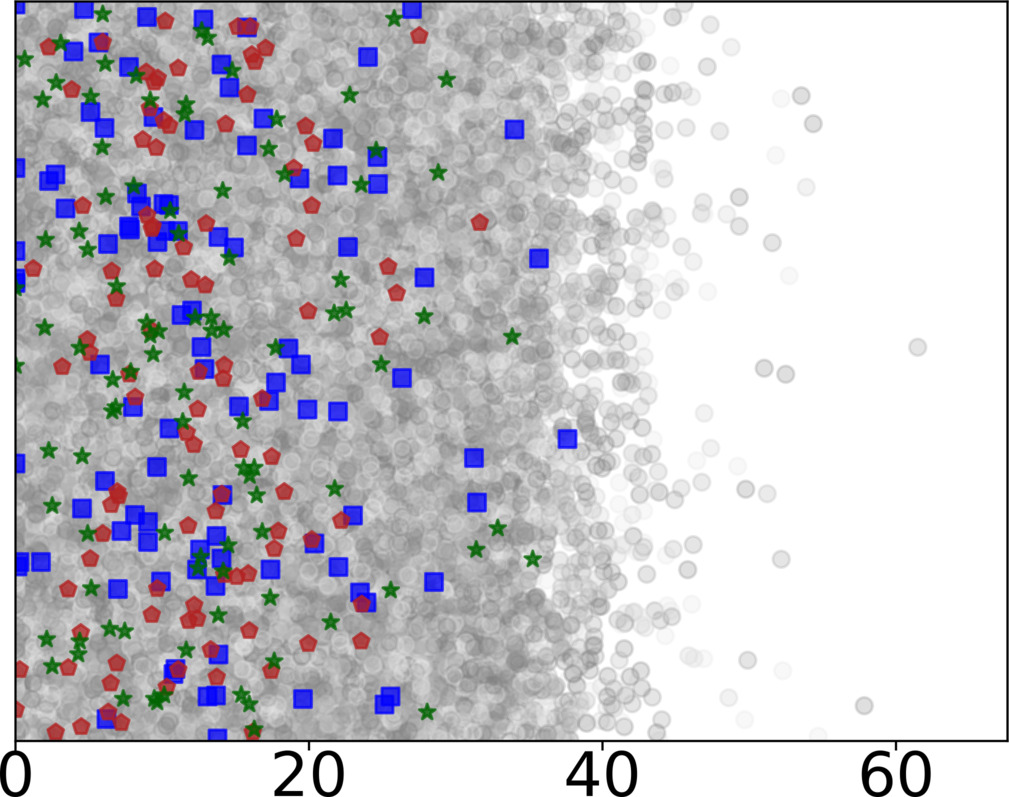} & 
    \includegraphics[width=0.32\linewidth]{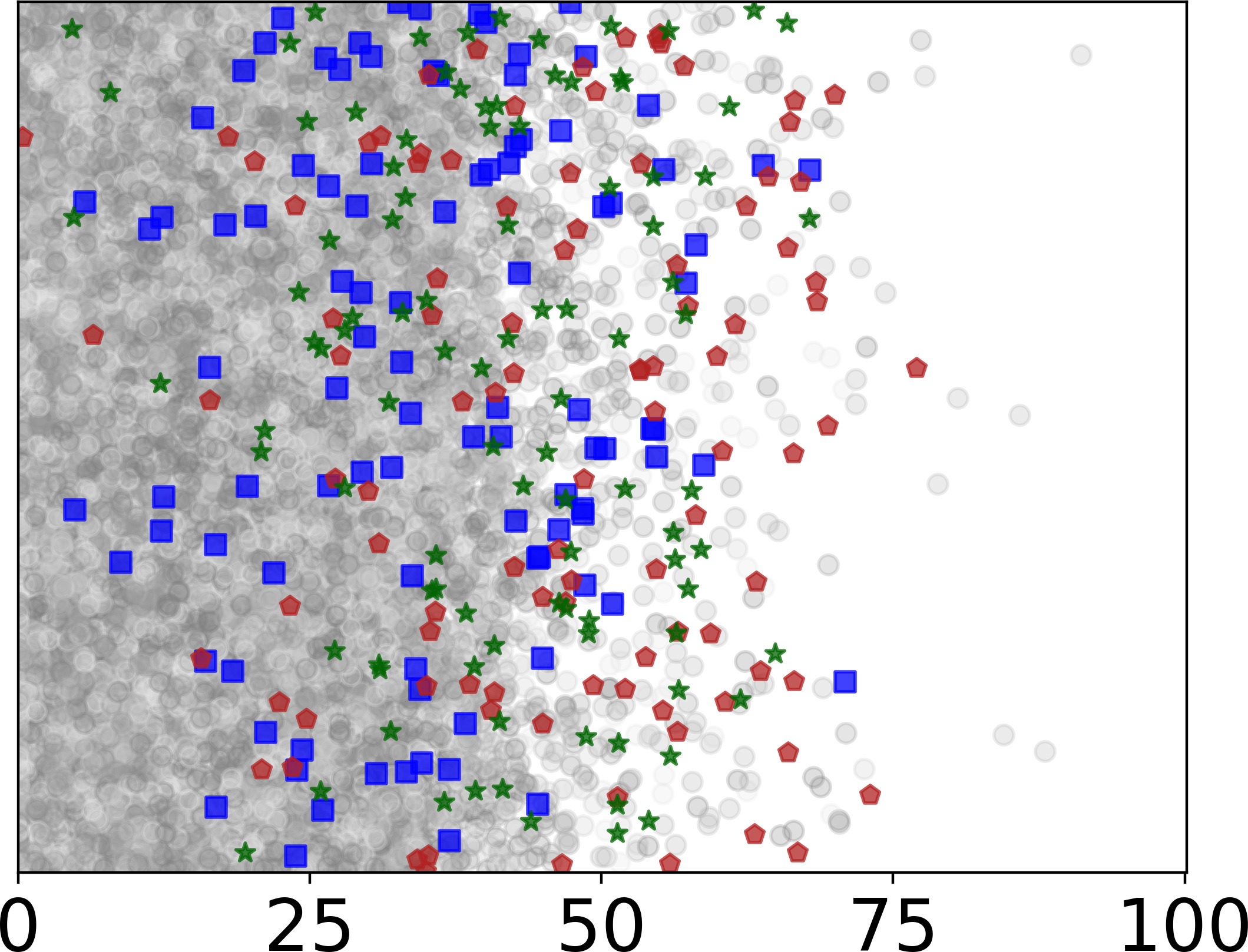} & 
    \includegraphics[width=0.305\linewidth]{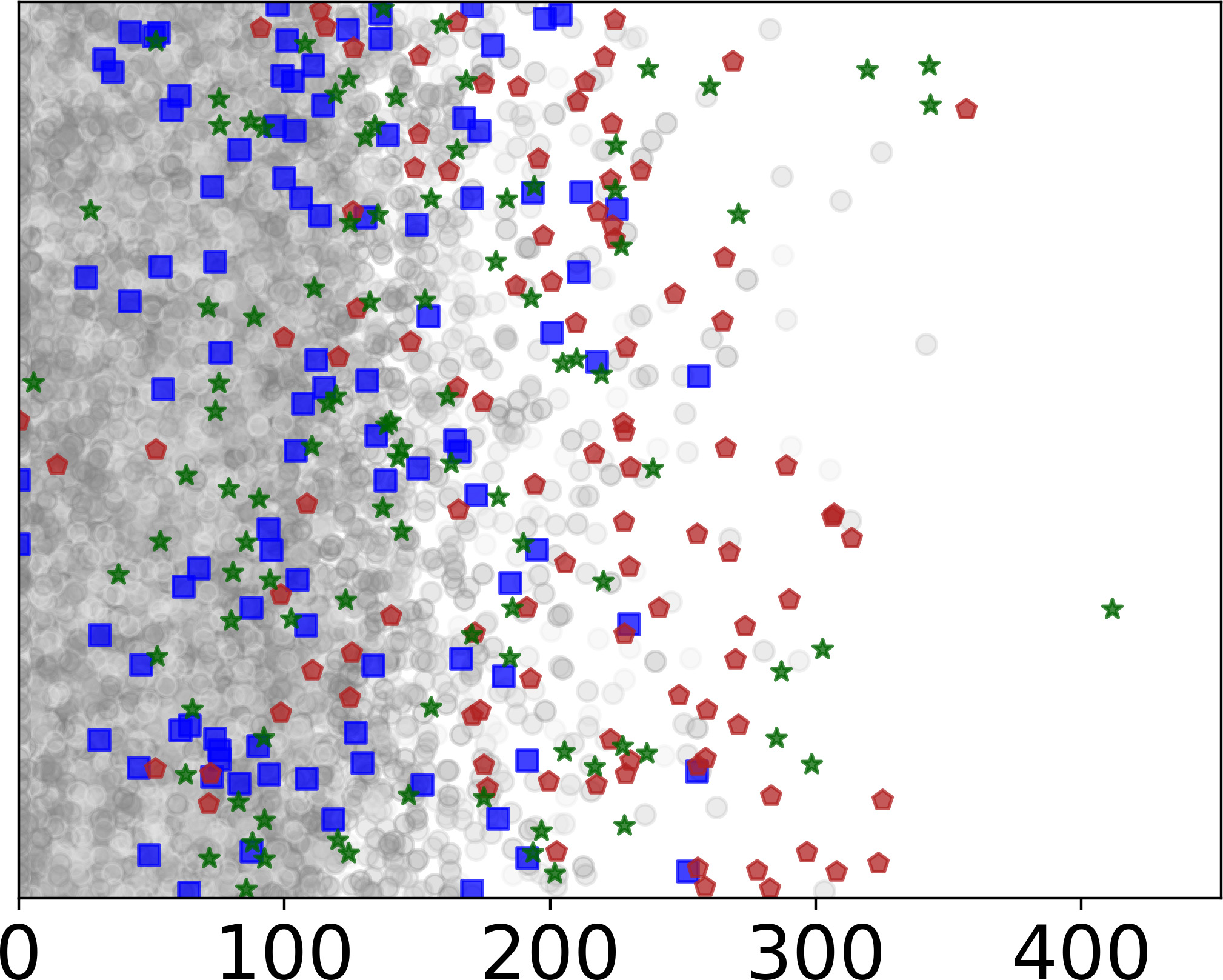}\\
    a. GoogLeNet (ImageNet) & b. GoogLeNet (Places-365)  & c. VGG-16 (Places-365)\\
    inception4e.494 & inception4e.824 & conv5\_3.20\\
    \precision: 0.0 & \precision: 0.27 & \precision: 0.53\\
    CCMAS:0.52 & CCMAS: 0.55 & CCMAS: 0.82\\
    $\mu_A$ = 72.5~~~$\mu_{\neg A}$ = 22.8
    & 
    $\mu_A$ = 41.0~~~$\mu_{\neg A}$ = 11.8 
    & $\mu_A$ = 157.6~~~$\mu_{\neg A}$ = 15.2\\
\end{tabular}
\caption{The units with with the highest Network Dissection scores for the category `bus'. The scatter plots, \precision, and CCMAS scores all indicate a low selectivity for this category. 
\blue squares: `school bus'; 
\red pentagons: `trolleybus'; \green stars: `minibus'; 
\grey circles: other classes.}
\label{fig:zhou}
\end{figure*}

Thus far we have assessed the selectivity of hidden units in AlexNet and shown that no units can reasonably be characterized as object detectors despite the high\precision and CCMAS scores of some units. This raises the question as to whether more recent CNNs learn object detector units.  In order to address this, we display jitterplots for three units that have the highest IoU scores according to the Network Dissection for the category BUS in (a) GoogLeNet trained on ImageNet, (b) GoogLeNet trained on Places-365, and (c) VGG-16 trained on Places-365, respectively \citep{102}, see figure~\ref{fig:zhou}. Models trained on the Places-365 dataset learn to categorize images into scenes (e.g., bedrooms, kitchens, etc.) rather than into object categories, and nevertheless, \citet{102} reported more object detectors in models trained on the Places-365 dataset (e.g., selective for objects within a scene such as a lamp in a bedroom or a car on a highway) than for ImageNet. We illustrate the selectivity of the BUS category because it corresponds to three output categories in ImageNet so we can easily plot the jitterplots for these units.

As was the case with AlexNet, the jitterplots show that the most selective units display some degree of selectivity, with the BUS images more active on average compared to non-Buses. 
However, these units are no more selective than the units we observed in AlexNet.  Indeed, the \precision measure of selectivity for the first units is 0.0, with none of the three units having a \precision of .75 that was the criterion of object detectors by \citet{105}, and CCMAS scores for first two units were roughly similar to the mean CCMAS score for AlexNet units in \layer{conv5}  (and much lower than the mean in \layer{fc6} and \layer{fc7} ). The most selective VGG-16 unit trained on Places-365 has lower \precision and CCMAS scores than the Monarch Butterfly unit depicted in Figure 3. So again, different measures of selectivity support different conclusions, and even the most selective units are far from the selective units observed in recurrent networks as reported in Figure 1a. See Tables \ref{tab:object_detVGG} - \ref{tab:object_detGNImNet} in Appendix D for more details about these and other units.

\section{Discussions and Conclusions}

Our central finding is that different measures of single-unit selectivity for objects support very different conclusions when applied to the same units in AlexNet.  In contrast with the \precision \citep{105} and CCMAS \citep{101} measures that suggest some highly selective units for objects in layers \layer{conv5}, \layer{fc6}, and \layer{fc7}, the recall with perfect \precision and false alarm rates at maximum informedness show low levels of selectivity. Indeed, the most selective units have a poor hit-rate or a high false-alarm rate (or both) for identifying an object class. The same outcome was observed with units in VGG-16 and GoogLeNet trained on either ImageNet or the Places-365 dataset. 

Not only do the different measures provide very different assessments of selectivity, the \precision, CCMAS, and Network Dissection measures provide misleading estimates of selectivity that have led to mistaken conclusions. For example, unit fc6.1199 in AlexNet trained on ImageNet is considered an Monarch Butterfly detector according to \citet{105} with a \precision score of .98 (and a CCMAS score of .93).  But the jitterplot in Fig. 3 and signal detection scores (e.g., high false alarm rate at maximum informedness) show this is a mischaracterisation of this unit. In the same way, the Network Dissection method identified many object detectors in VGG-16 and GoogLeNet CNNs, but the jitterplots in Fig.~\ref{fig:zhou}  show that this conclusion is unjustified. 

What level of selectivity is required before a unit can be considered an `object detector' for a given category?  In the end, this is a terminological point. On an extreme view, one might limit the term to the `grandmother units' that categorize objects with perfect recall and specificity, or alternatively, it might seem reasonable to describe a unit as a detector for a specific object category if there is some threshold of activation that supports more hits than misses (the unit is strongly activated by the majority of images from a given category), and at the same time, supports more hits than false alarms (the unit is strongly activated by items from the given category more often than by items from other categories). Or perhaps a lower standard could be defended, but in our view, the term `object detector' suggests a higher level of selectivity than 8\% recall at perfect precision. That said, our results show that some units respond strongly to some (unknown) features that are weakly correlated with an object category.  For instance, unit fc6.1199 is responding to features that occur more frequently in Monarch Butterflies than other categories.  This can also be seen in a recent ablation study in which removing the most selective units tended to impair the CNN's performance in identifying the corresponding object categories more than other categories \citep{zhou2018revisiting}.  But again, the pattern of performance is not consistent with the units being labeled `object detectors'. 

What should be made of the finding that localist representations are sometimes learned in RNNs (units with perfect specificity and recall), but not in AlexNet and related CNNs? The failure to observe localist units in the hidden layers of these CNNs is consistent with \citet{1}'s claim that these units emerge in order to support the co-activation of multiple items at the same time in short-term memory. That is, localist representations may be the solution to the superposition catastrophe, and these CNNs only have to identify one image at a time. The pressure to learn highly selective representations in response to the superposition constraint may help explain the reports of highly selective neurons in cortex given that the cortex needs to co-activate multiple items at the same time in order to support short-term memory~\citep{6}. 

At the same time, it should be emphasized that the RNNs that learned localist units were very small in scale compared to CNNs we have studied here, and accordingly, it is possible that the contrasting results reflect the size of the networks rather than the superposition catastrophe \textit{per se}. Relevant to this issue a number of authors have reported the existence of selective units in larger RNNs with long-short term memory (LSTM) units \citep{kaparthy2016visualizing, radford2017learning, lakretz2019emergence,na2018discovery}.  Indeed, \citet{lakretz2019emergence} use the term `grandmother cell' to describe the units they observed. It will be interesting to apply our measures of selectivity to these larger RNNs and see whether these units are indeed `grandmother units'. It should also be noted that there are recent reports of impressively selective representations in generative adversarial networks \citep{bau2019visualizing} and variational autoencoders \citep{burgess2018understanding} where the superposition catastrophe is not an issue. Again, it will be interesting to assess the selectivity of these units according to signal detection measures in order to see whether there are additional computational pressures to learn highly selective or even grandmother cells. We will be exploring these issues in future work.

Finally, how do these findings relate to the selectivity of neurons in visual cortex? Is the limited degree of selectivity observed in various CNNs a problem for the claim that these models provide a good theory of human vision?  Not necessarily.  As noted at the start, there is currently a debate about how selective neurons are in cortex, and few researchers have carried out relevant behavioral experiments that can be compared to the selectivity studies carried out in CNNs \citep{Bowers2019}. Furthermore, there is widespread confusion about what constitutes a localist grandmother cell, with some versions of localist units responding to multiple different categories, with one category more active than all others. \citep{2}. This makes it all the more challenging to determine whether the human visual system learns to identify objects on the basis of localist codes.  Nevertheless, a better understanding of the selectivity of units in CNNs and other artificial networks is a necessary step towards a better understanding of the relation between these models and human visual system. And adopting a standard set of measures will allow researchers to compare selectivity across different network architectures trained on different tasks in order to better understand the factors that contribute to more or less selectivity.

\section*{Acknowledgments}
This project has received funding from the Leverhulme Trust (grant no. RPG-2016-113) and the European Research Council (ERC) under the European Union’s Horizon 2020 research and innovation programme (grant agreement No. 741134). AN was supported by the National Science Foundation CRII grant No. 1850117.

\clearpage
\bibliography{ref2}


\appendix

\begin{center}
\Large 
Appendix
\end{center}

\setcounter{table}{0}
\setcounter{figure}{0}
\renewcommand\thetable{A\arabic{table}}
\renewcommand\thefigure{A\arabic{figure}}

\section{Instructions for behavioral experiment}

Participants were provided the following instructions: "In each trial, a grid of computer generated images will be presented which may have recognisable common objects (e.g. car, trash can, banjo, clothes), animals (fish, bird, dog), or places (theatre, viaduct, volcano). In each trial, you will be asked whether you can identify multiple examples of an everyday object, place or animal." The full set of instructions, together with all of the stimuli used in the task are stored here: \url{https://gorilla.sc/openmaterials/84689} and more details of the task are described under `Methodological details for the behavioral experiment' in Section \ref{sec:psychometh} of the manuscript.  Below is a screen shot from an example trial where they were asked three questions of actual images.

\begin{figure}[bh]
\includegraphics[width=\textwidth]{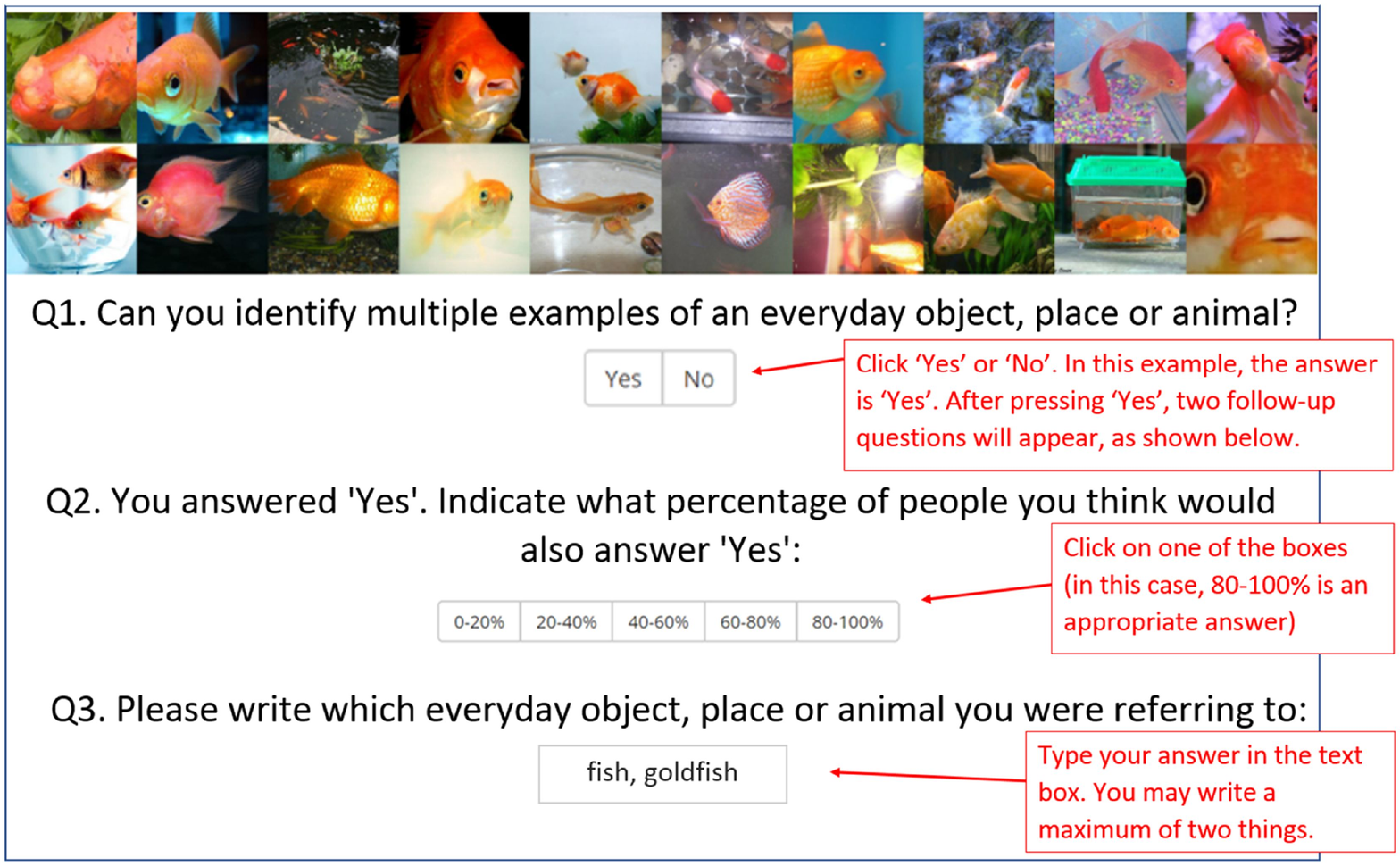}
\caption{Example screen from the identification task shown to participants as part of the instructions. The images included on this practice trial are ImageNet2012 images, not AM images.}
\label{fig:taskZb}
\end{figure}

\section{Further data on the selectivity measures across AlexNet}

Table ~\ref{tab:XTream} gives the highest values of CCMAS and \precision for each layer in AlexNet, with the corresponding CCCMAS and \precision scores for these units. It is worth noting that the highest CCMAS score of all hidden units was .94 (\layer{fc7}.\unit{31}), which at first glance suggests that this unit is close to `perfect' selectivity.  However, this unit only has low a \precision score of .11. (Note: \precision in this analysis used the 100 most active items, rather than the 60 most active items).  In other words, although the mean activation for the given class is very high relative to the mean of all other activations (high CCMAS), the proportion of items from that class in the 100 most active items is low (low \precision).  See~\ref{sec:CCMAS2} for some discussion of how this might occur.  

\begin{table}[htb]
\begin{center}
\begin{tabular}{lll}
\multicolumn{1}{c}{\bf LAYER.UNIT} & \multicolumn{1}{c}{\bf CCMAS} & \multicolumn{1}{c}{\bf Precision}
\\\hline
\\\multicolumn{3}{c}{\bf Top CCMAS units}\\
\layer{output}.\unit{322}		& 0.991 	& 1.0	\\
\layer{fc7}.\unit{31}		& 0.94	& .11	\\
\layer{fc6}.\unit{582}		& 0.93	& .01 	\\
\layer{conv5}.\unit{78}		& 0.75 	& .05	\\
\\ \multicolumn{3}{c}{\bf Top \precision units}\\
\layer{output}.\unit{0}		& 0.99 & 1.0	\\
\layer{fc7}.\unit{255}		& 0.90	& .97	\\
\layer{fc6}.\unit{1199}		& 0.92	& .95	\\
\layer{conv5}.\unit{0}		& 0.55 & .77	\\ 
\end{tabular}
\caption{The units with the highest CCMAS and \precision scores in AlexNet.  Unit \layer{fc6}.\unit{1199} was displayed in Fig. \ref{fig:butterfly}.}
\label{tab:XTream}
\end{center}
\end{table}


Table ~\ref{tab:correlations} shows positive correlations between four of the selectivity measures used.  There are moderate positive correlations between \precision and CCMAS; and between \precision and Recall at 95\% precision.  The other correlations between selectivity measures have weak positive correlations.  All four selectivity measures are negatively correlated with the number of classes present in the 100 most active items, that is, the more selective the unit, the fewer classes will be represented in the most active 100 items.



\begin{table}[h]
    \centering
    \begin{tabular}{c|cccc}
    & CCMAS & recall|$_{0.95}$  & Max. Inf. & No. classes in top100\\
    \hline
       \precision  & 0.38 & 0.30 & 0.15 & -0.68 \\
       CCMAS &  & 0.09 & 0.14 & -0.47 \\ 
       recall|$_{0.95}$  & & & 0.10  & -0.19 \\
       Max. Inf. & & & & -0.22 \\
    \end{tabular}
    \caption{The correlations between the different measures. (All \textit{p}'s < .001)}
    \label{tab:correlations}
\end{table}

\section{Further issues with the CCMAS measure\label{sec:CCMAS2}}



The CCMAS measure is based on comparing the mean activation of a category with the mean activation for all other items, and this is problematic for a few reasons.  First, in many units a large proportion of images do not activate a unit at all.  For instance, our butterfly `detector' unit \layer{fc6}.\unit{1199} has a high proportion of images with an activation of 0.0 (see figure ~\ref{fig:butterfly}).  Indeed, the inset on the middle figure shows that the distribution can be better described by exponential-derived fits rather than a Gaussian.  This means that the CCMAS selectivity is heavily influenced by the the proportion of images that have an activation value of zero (or close to zero).  This can lead to very different estimates of selectivity for CCMAS and \precision or localist selectivity, which are driven by the most highly activated items.  

\begin{figure*}[htb]
	\begin{subfigure}{0.3\linewidth}
		\centering		
		\includegraphics[width=1.0\linewidth]{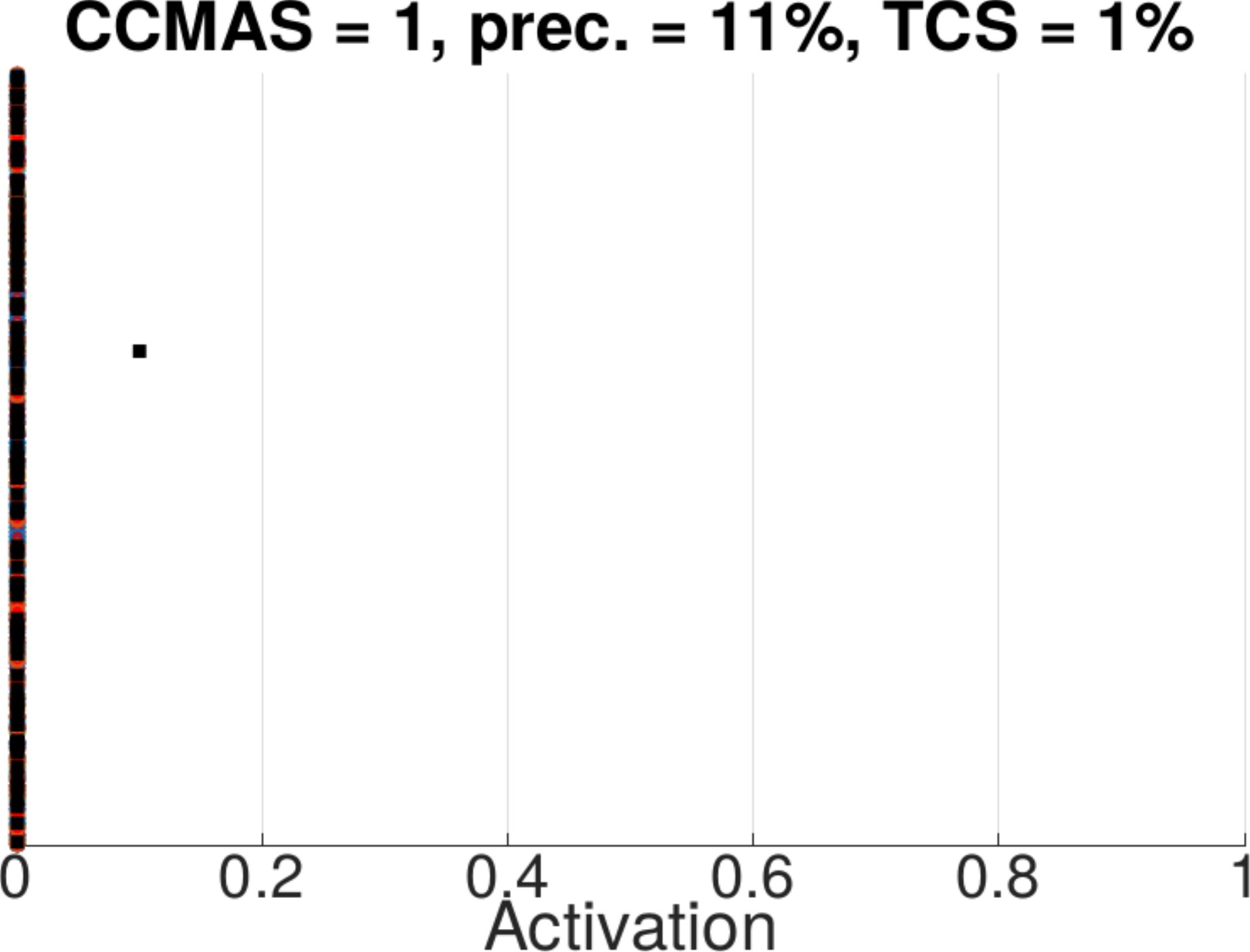}
		\caption{One active item from one class. \\CCMAS = 1\\\precision = .11}
	\end{subfigure}	
	~
	\begin{subfigure}{0.3\linewidth}
		\centering
		\includegraphics[width=1.0\linewidth]{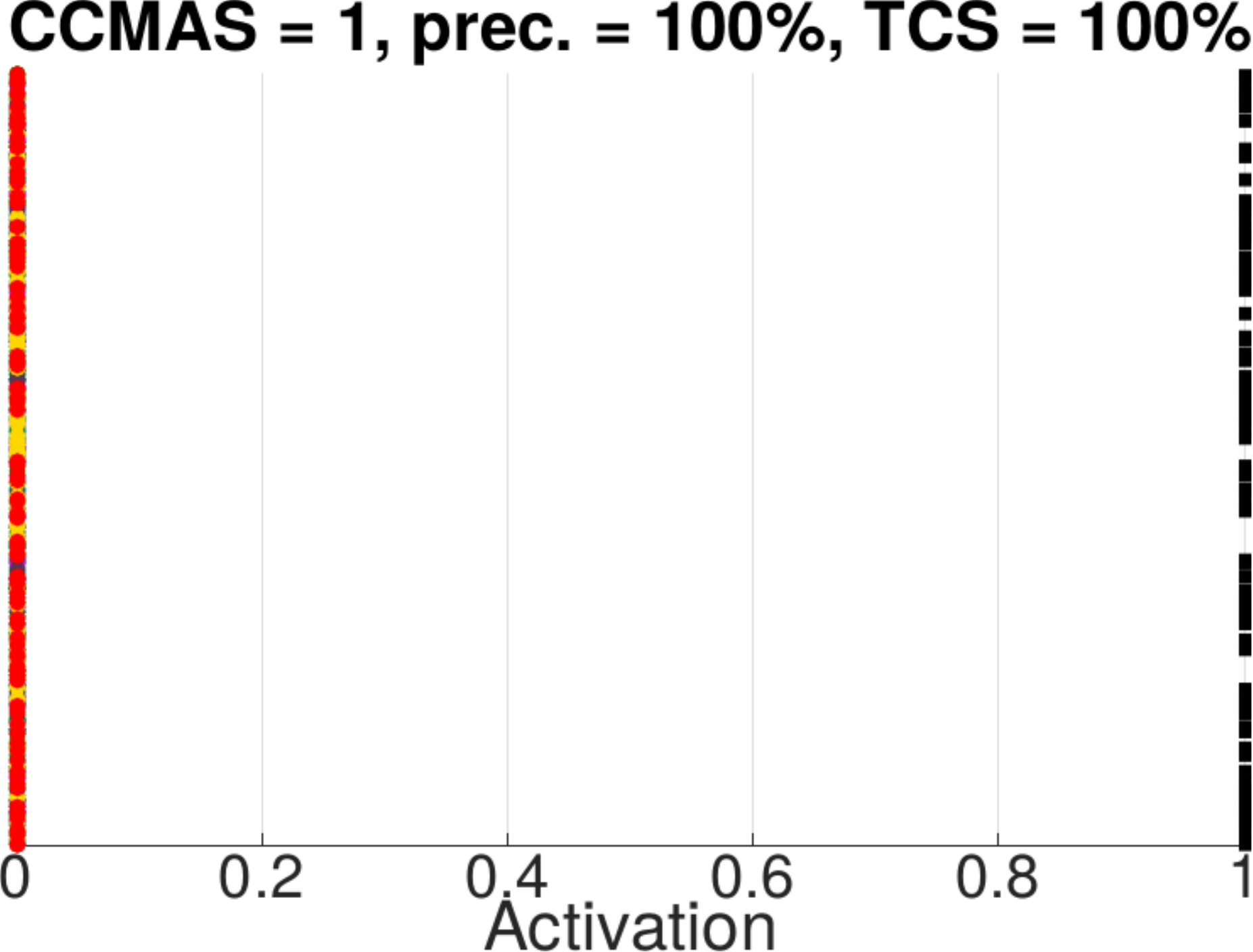}
		\caption{Archetypal `GMC' unit.\\CCMAS = 1\\\precision = 1.0}
	\end{subfigure}	
	~
    \begin{subfigure}{0.31\linewidth}
		\centering
		\includegraphics[width=1.0\linewidth]{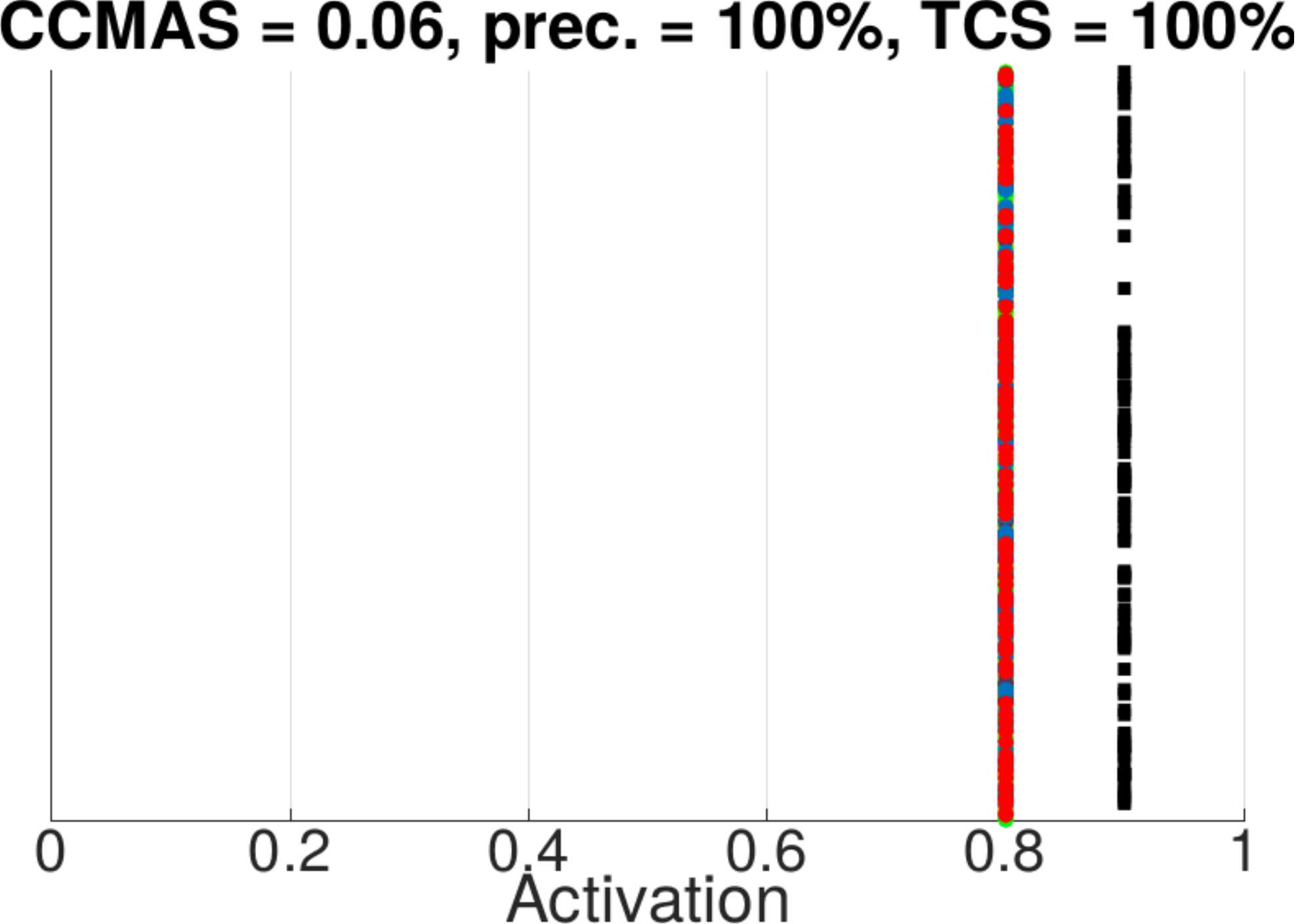}
		\caption{One class more active than the others.\\CCMAS = 0.06 \\\precision = 1.0}	
	\end{subfigure}	
    	\hspace{1mm}
    \caption{Example of where the CCMAS does not match intuitive understandings of selectivity. Generated example data: (a) If a unit is off to all but a single image from a large class of objects, the CCMAS for that class is 1 (maximum possible selectivity). (b) An archetypal `grandmother' cell (GMC), where the unit is strongly activated to all members of one class and off to everything else.  The CCMAS is the same for (b) as for (a) although the \precision is vastly different. (c): If a unit has high activations for all classes, but one class (black squares) is 0.1 more than all others (coloured circles), the CCMAS is very low (0.06) despite being 1.0 precision. The generated examples are for 10 classes of 100 items}
\label{fig:CCMASzBs}
\end{figure*}

In Figure ~\ref{fig:CCMASzBs} we generate example data to highlight ways in which CCMAS score may be non-intuitive.  In subplot (a) we demonstrate that a unit can have a CCMAS score of of 1.0 despite only a single item activating the unit.  The point that we wish to emphasise is that a high CCMAS score does not necessarily imply selectivity for a given class, but might in fact relate to selectivity for a small subset of items from a given class, and this is especially true when a unit's activation is sparse (many items do not activate the unit).  However, the reverse can also be true.  In subplot (c) we demonstrate that a unit can have a very low CCMAS score of .06 despite all of the most active items being from the same class.

In addition, if the CCMAS provided a good measure of a unit’s class selectivity, then one should expect that a high measure of selectivity for one class would imply that the unit is not highly selective for other classes. However, the CCMAS score for the most selective category and the second most selective category (CCMAS2) were similar across the \layer{conv5} , \layer{fc6} and \layer{fc7} . layers, with the mean CCMAS scores .491, .844, and .848, and the CCMAS2 scores .464, .821, .831. For example, unit \layer{fc7} .0 has a CCMAS of .813 for the class ‘maypole’, and a CCMAS2 score of .808 for ‘chainsaw’ (with neither of these categories corresponding ‘orangutan’ that had the highest \precision of score of .14).



\section{Testing units in other models}
\label{sec:test_other_models}

To investigate units characterized by \citet{102} to be object detectors, 
we focus on units from a single layer that are reported to be `bus detectors', that is, units with an IoU $\geq .04$.  We used the first 100 images per class from the ImageNet 2012 dataset as our test data. There are three classes of bus in this dataset: `n04146614 school bus', `n04487081 trolleybus, trolley coach, trackless trolley', `n03769881 minibus', and this corresponded to 300 items out of 100000 images. 
Data for all bus unit detectors for VGG trained on places 365 are shown in table~\ref{tab:object_detVGG}; for GoogLeNet trained on places 365 in table~\ref{tab:object_detGNPlaces}; and for GoogLeNet trained on ImageNet are shown in table~\ref{tab:object_detGNImNet}.  Note that for all units there are very few busses with activation at zero and that the mean activation for busses is higher than the mean activation for non-busses.  However, all \precision scores are all below .6, meaning that of the 100 items that most strongly activated the unit, at least 40 of them were not busses. Together these results suggests that whilst these units demonstrate some sensitivity to busses, they show poor specificity for busses (e.g., high false-alarm rate).  



\begin{table}[htb]
    \centering
    \begin{tabular}{c|cc|cc|cc|cc}
        unit & IoU & top-4 match & no.$a_x$>0 & no.$a_x$>0 & $\mu_A$ & $\mu_{\neg A}$ & \precision & CCMAS \\
        &&& $x \in A$ & $x \in \neg A$ &&&\\
        \hline
         \small \layer{conv5\_3}$_\unit{191}$  & .15 & Y & 99.0\% & 63.9\% & 131.9 & 16.1 & .45 & .78\\
         \small \layer{conv5\_3}$_\unit{20}$  & .15 & Y & 99.0\% & 49.1\% &  157.6 & 15.2 & .53 & .82\\
         \small \layer{conv5\_3}$_\unit{333}$  & .08 & Y & 99.0\% &  71.4\% &  101.7 & 17.5 & .24 & .71\\
        \small \layer{conv5\_3}$_\unit{145}$  & .07 & Y & 97.3\% & 61.7\% &  75.5 & 12.5 & .19 & .72\\
        \small \layer{conv5\_3}$_\unit{113}$  & .06 & N & 97.4\% & 41.0\% &  62.8 & 9.1 & .12 & .75\\
        \small \layer{conv5\_3}$_\unit{443}$  & .04 & N & 95.3\% & 38.2\% &  59.3 & 8.1 & .12& .76\\
        \small \layer{conv5\_3}$_\unit{131}$  & .04 & N &  93.7\% & 22.3\% &  54.0 & 5.86 & .08 & .80\\
    \end{tabular}
    \caption{Selectivity measures for VGG-16, trained on Places-365, top convolutional layer units identified by \citet{102} as object detectors.  Standard errors not shown for space, but were below $\pm 5$.  The IoU is from \citet{102}'s network dissection method. no.$a_x$>0 and no.$a_x$>0 $x \in A$ refer to the proportion of activations that were greater than zero for busses and non-busses respectively.  $\mu_A$ and $\mu_{\neg A}$ are the class means for busses and non busses respectively. A unit was coded as top-4 match (Y) if there was a single bus in the 4 example pictures on the website (\url{http://netdissect.csail.mit.edu/dissect/vgg16_places365/}), and (N) if not.}
    \label{tab:object_detVGG}
\end{table}


\begin{table}[htb]
    \centering
    \begin{tabular}{c|cc|cc|cc|cc}
        unit & IoU & top-4 match &  no.$a_x$>0 & no.$a_x$>0 & $\mu_A$ & $\mu_{\neg A}$ & \precision & CCMAS \\
        &&& $x \in A$ & $x \in \neg A$ &&&&\\
        \hline
        $_\unit{824}$  & .17 & Y &
         100.0 & 91.4 & 41.0 & 11.8 & .27 & .55 \\
        $_\unit{745}$  & .13 & Y &
        98.3 & 74.8 & 34.8 & 11.4 & .06 & .51\\
        $_\unit{791}$  & .11 & Y &
        98.3 & 71.4 & 32.7 & 5.3 & .41 & .72\\
        $_\unit{194}$  & .11 & N &
        100.0 & 85.3 & 26.6 & 8.8 & .02 & .51 \\
        $_\unit{82}$  & .11 & Y &
        100.0 & 97.3 & 26.7 & 10.9 & .14 & .42\\
        $_\unit{736}$  & .11 & N &
         100.0 & 78.8 & 38.7 & 9.9 & .05 & .59\\
        $_\unit{663}$  & .10 & N &
        96.0 & 38.0 & 33.4 & 3.7 & .15 & .80\\
        $_\unit{94}$  & .10 & Y &
        100.0 & 91.6 & 38.3 & 9.5 & .35 & .60\\
        $_\unit{772}$  & .08 & N &
        97.3 & 54.6 & 21.7 & 5.2 & .00 & .61\\
        $_\unit{113}$  & .08 & N &
        100.0 & 88.0 & 24.9 & 9.2 & .02 & .46\\
        $_\unit{708}$  & .06 & N &
        100.0 & 85.1 & 29.7 & 9.1 & .02 & .53\\
        $_\unit{801}$  & .06 & N &
        100.0 & 64.5 & 35.2 & 6.4 & .14 & .69\\
        $_\unit{199}$  & .06 & N &
        99.7 & 92.2 & 21.5 & 7.7 & .09 & .47\\
        $_\unit{8}$  & .05 & N &
        99.7 & 83.5 & 18.5 & 7.3 & .01 & .43\\
        $_\unit{121}$  & .05 & N &
        100.0 & 90.4 & 17.9 & 8.9 & .01 & .34\\
        $_\unit{622}$  & .05 & Y &
        96.0 & 65.0 & 27.5 & 6.4 & .20 & .62\\
        $_\unit{97}$  & .04 & Y &
        99.3 & 86.4 & 21.1 & 9.3 & .04 & .39\\
    \end{tabular}
\caption{Selectivity measures for GoogLeNet, trained on Places-365, layer \layer{inception4e} units identified by \citet{102} as object detectors. Standard errors not shown for space, but were below $\pm 2$. The IoU is from \citet{102}'s network dissection method. A unit was coded as top-4 match (Y) if there was a single bus in the 4 example pictures on the website (\url{http://netdissect.csail.mit.edu/dissect/googlenet_places365/}), and (N) if not. no.$a_x$>0 and no.$a_x$>0 $x \in A$ refer to the proportion of activations that were greater than zero for busses and non-busses respectively.  $\mu_A$ and $\mu_{\neg A}$ are the class means for busses and non busses respectively.}
\label{tab:object_detGNPlaces}
\end{table}

\begin{table}[htb]
    \centering
    \begin{tabular}{c|cc|cc|cc|cc}
        unit & IoU & top-4 match &  no.$a_x$>0 & no.$a_x$>0 & $\mu_A$ & $\mu_{\neg A}$ & \precision & CCMAS \\
        &&& $x \in A$ & $x \in \neg A$ &&&&\\
        \hline
        $_\unit{494}$  & .11 & N &
        99.0 & 82.4 & 72.5 & 22.8 & .00 & .52\\
        $_\unit{828}$  & .10 & Y &
        100.0 & 72.6 & 109.4 & 17.6 & .45 & .72\\
        $_\unit{569}$  & .10 & Y &
        99.7 & 85.9 & 74.9 & 20.0 & .05 & .58\\
        $_\unit{384}$  & .10 & Y &
        100.0 & 71.6 & 67.0 & 18.5 & .00 & .57\\
        $_\unit{455}$  & .09 & Y &
        99.7 & 89.6 & 69.1 & 14.3 & .3 & .66\\
        $_\unit{579}$  & .09 & Y & 
        100.0 & 97.0 & 91.5 & 26.0 & .23 & .56\\
        $_\unit{331}$  & .08 & Y &
        98.0 & 75.5 & 51.0 & 11.8 & .12 & .62\\
        $_\unit{582}$  & .08 & Y &
        100.0 & 83.4 & 125.7 & 21.95 & .58 & .70\\
        $_\unit{498}$  & .07 & Y &
        97.7 & 77.2 & 73.5 & 15.0 & .52 & .66\\
        $_\unit{534}$  & .07 & N & 99.3 & 81.2 & 62.7 & 19. & .02 & .53\\
        $_\unit{693}$  & .07 & Y & 
        98.7 & 91.2 & 75.4 & 22.3 & .15 & .54\\
        $_\unit{673}$  & .07 & Y & 
        99.7 & 88.4 & 88.6 & 23.0 & .33 & .59\\
        $_\unit{469}$  & .06 & Y & 
        98.7 & 78.1 & 34.7 & 14.6 & .00 & .41\\
        $_\unit{207}$  & .06 & Y & 100.0 & 93.5 & 76.1 & 21.3 & .07 & .56\\
        $_\unit{491}$  & .06 & N &
        99.0 & 74.5 & 41.1 & 13.7 & .01 & .50\\
        $_\unit{645}$  & .06 & Y &
        98.0 & 83.9& 59.9 & 18.1 & .20 & .54\\
        $_\unit{527}$  & .06 & N &
        100.0 & 91.5 & 58.0 & 21.7 & .00 & .46\\
        $_\unit{511}$  & .05 & N &
        100.0 & 89.4 & 53.5 & 21.7 & .00 & .42\\
        $_\unit{308}$  & .05 & N &
        100.0 & 89.4 & 53.5 & 21.7 & .00 & .42\\
        $_\unit{541}$  & .05 & N &
        99.67 & 88.7 & 44.9 & 13.7 & .00 & .53\\
        $_\unit{367}$  & .05 & Y &
        97.3 & 80.3 & 37.7 & 15.4 & .02 & .42\\
        $_\unit{665}$  & .05 & Y &
        100.0 & 82.45 & 107.2 & 21.0 & .33 & .67\\
        $_\unit{532}$  & .05 & Y &
        100.0 & 91.5 & 52.9 & 22.4 & .05 & .41\\
        $_\unit{297}$  & .04 & Y &
        99.7 & 90.2 & 48.2 & 17.9 & .00 & .46\\
        $_\unit{480}$  & .04 & Y &
        100.0 & 92.9 & 69.4 & 21.4 & .02 & .53\\

    \end{tabular}
\caption{Selectivity measures for GoogLeNet, trained on ImageNet, layer \layer{inception4e} units identified by \citet{102} as object detectors. Standard errors not shown for space, but were below $\pm 2$. A units is marked as a top-4 match (Y) if there was a single bus in the 4 example pictures on the website (\url{http://netdissect.csail.mit.edu/dissect/googlenet_imagenet/}), and (N) if not. }
\label{tab:object_detGNImNet}
\end{table}

\end{document}